\documentclass[runningheads]{llncs}
\usepackage{graphicx}

\usepackage{tikz}
\usepackage{comment}
\usepackage{amsmath,amssymb} 
\usepackage{color}
\usepackage{graphicx}
\usepackage{amsmath}
\usepackage{amssymb}
\usepackage{booktabs}
\usepackage{hyperref}
\usepackage{enumitem}
\usepackage{times}
\usepackage{epsfig}
\usepackage{multirow}
\usepackage{mathtools}
\usepackage{cuted}
\usepackage{capt-of}
\usepackage[ruled,vlined]{algorithm2e}
\usepackage[normalem]{ulem}
\useunder{\uline}{\ul}{}

\usepackage[accsupp]{axessibility}  

\def\etal{\emph{et al.}}

\def\ie{\emph{i.e.}}

\usepackage[capitalize]{cleveref}
\crefname{section}{Sec.}{Secs.}
\Crefname{section}{Section}{Sections}
\Crefname{table}{Table}{Tables}
\crefname{table}{Tab.}{Tabs.}

\begin{document}
\pagestyle{headings}
\mainmatter

\title{3SD: Self-Supervised Saliency Detection With No Labels} 



\titlerunning{3SD}
\author{Rajeev Yasarla\inst{1}\and
Renliang Weng\inst{2} \and
Wongun Choi\inst{2} \and
Vishal Patel\inst{1}\and
Amir Sadeghian\inst{2}}
\authorrunning{R. Yasarla et al.}
\institute{Johns Hopkins University, Baltimore, MD 21218, USA
\email{\{ryasarl1,vpatel36\}@jhu.edu}\\ \and
AIBEE Inc., Palo Alto, CA 94306, USA\\
\email{\{rlweng,wgchoi,amirabs\}@aibee.com}}

\maketitle

\begin{abstract}

We present a conceptually simple  self-supervised method for saliency detection. Our method generates and uses pseudo-ground truth labels for training. The generated pseudo-GT labels don't require any kind of human annotations (\emph{e.g.}, pixel-wise labels or weak labels like scribbles).
Recent works show that features extracted from classification tasks provide important saliency cues like structure and semantic information of salient objects in the image. Our method, called 3SD, exploits this idea by adding a branch for a self-supervised classification task in parallel with salient object detection, to obtain class activation maps (CAM maps). These CAM maps along with the edges of the input image are used to generate the pseudo-GT saliency maps to train our 3SD network. Specifically, we propose a contrastive learning-based training on multiple image patches for the classification task. We show the multi-patch classification with contrastive loss improves the quality of the CAM maps compared to naive classification on the entire image. Experiments on six benchmark datasets demonstrate that without any labels, our 3SD method outperforms all existing weakly supervised and unsupervised methods, and its performance is on par with the fully-supervised methods. Code is available at :\textit{\href{https://github.com/rajeevyasarla/3SD}{https://github.com/rajeevyasarla/3SD}}

\end{abstract}
\begin{figure}[h!]
    \centering
	\includegraphics[width=0.95\linewidth]{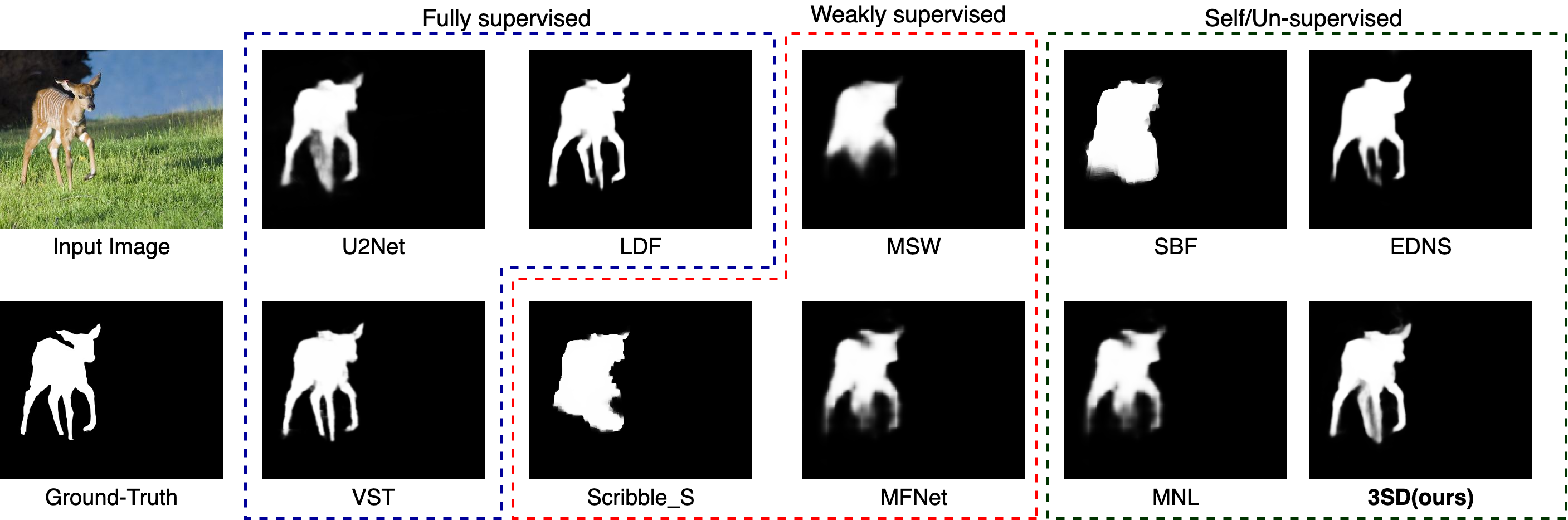} 
	\vskip -5pt 
	\caption{Comparisons of saliency maps. Self-supervised: \textbf{3SD (ours)}. Weakly supervised:  MSW\cite{zeng2019multi}, Scribble\cite{zhang2020weakly}, MFNet\cite{Piao_2021_ICCV}. Un-supervised: SBF\cite{zhang2017supervision},  MNL\cite{zhang2018deep}, EDNS\cite{zhang2020learning}. Fully-supervised:~U2Net\cite{qin2020u2}, LDF\cite{CVPR2020_LDF}, VST\cite{liu2021visual}. 
	}
	\label{Fig:fig1}
	\vspace{-2em}
\end{figure}
\vspace{-1em}
\section{Introduction}
Salient object detection (SOD) task is defined as pixel-wise segmentation of interesting regions that capture human attention in an image. It is widely used as a prior to improve many computer vision tasks such as visual tracking, segmentation, \emph{etc}. Early methods based on hand-crafted features like histograms \cite{lu2013robust}, boundary connectivity \cite{zhu2014saliency}, high-dimensional color transforms\cite{kim2014salient}, may fail in producing high-quality saliency maps on cluttered images where the foreground object is similar to the background. In recent years, deep convolutional neural networks (CNNs), and in particular fully convolutional networks (FCN)\cite{long2015fully} have provided excellent image segmentation and salient object detection performance. 

In general, the CNN-based salient object detection methods can be classified into three groups: (i) fully-supervised methods (that require large-scale datasets with pixel-wise annotations), (ii) weakly supervised, and (iii) unsupervised or self-supervised methods (that don't require actual pixel-wise annotations of salient object detection). The main drawback of the fully-supervised methods \cite{Qin_2019_CVPR,wu2019cascaded,qin2020u2,CVPR2020_LDF,liu2021visual} is that they require a large amount of pixel-wise annotations of salient objects which is time-consuming and expensive. On the other hand, to minimize human efforts in labeling datasets, weakly-supervised approaches \cite{li2018weakly,zeng2019multi}  have been proposed which address saliency detection either by using weak sparse labels such as image class labels~\cite{li2018weakly} or image captions~\cite{zeng2019multi}. 
Alternatively, Zhang~\etal~\cite{zhang2020weakly} present a weakly-supervised SOD method based on scribble annotations. Pia~\etal\cite{Piao_2021_ICCV} propose a learnable directive filter based method that extracts saliency cues using multiple labels from the attentions. Note, Pia~\etal\cite{Piao_2021_ICCV} and MSW~\cite{zeng2019multi} rely on the features or attention maps obtained from a classification task, and might fail to produce high-quality pseudo-GTs since the classification task is trained with global class label or image caption. For example, Fig. \ref{Fig:fig1} shows that the outputs of \cite{zeng2019multi,Piao_2021_ICCV} are not sharp and miss fine details like legs and ears. Although these weakly-supervised methods reduce the amount of labeling required for SOD, they still require labeling resources to obtain image captions~\cite{zeng2019multi}, image class labels~\cite{li2018weakly,Piao_2021_ICCV}, or accurate scribble annotations\cite{zhang2020weakly}. 
On the other hand, unsupervised methods \cite{zhang2018deep,nguyen2019deepusps,zhang2020learning} devise a refinement procedure or generative based approach (noise-aware), that utilizes the hand crafted features, and/or noisy annotations. Note that performance of these unsupervised methods highly rely on the noisy annotation, and might struggle to produce high-quality saliency maps if they fail to recover the underlying semantics from the noisy annotations. For instance, we can observe in Fig. \ref{Fig:fig1} saliency outputs of  \cite{zhang2018deep,nguyen2019deepusps,zhang2020learning} miss parts like legs and ears.

In an attempt to overcome these issues, we propose our Self-Supervised Saliency Detection (3SD) method. Our framework follows the conventional encoder-decoder structure to generate saliency map. In terms of encoder, we present a novel encoder architecture which consists of a local encoder and global context encoder. The local encoder learns pixel-wise relationship among neighbourhood while the global encoder encodes the global context. The outputs of these encoders are concatenated and subsequently fed to the decoder stream. By fusing both local features and global context we are able to extract both fine-grain contour details as well as adhere to the underlying object structure.

\begin{figure}[ht!]
		\centering
		
		\includegraphics[width=0.17\textwidth]{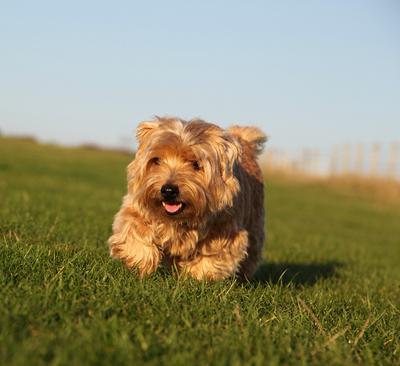}
		\includegraphics[width=0.17\textwidth]{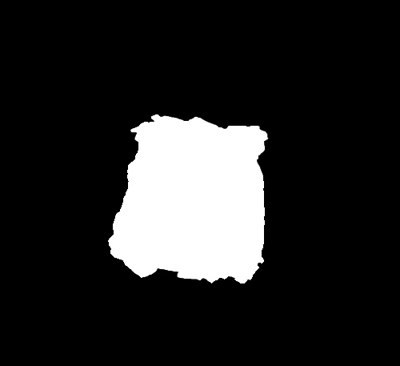}
		\includegraphics[width=0.17\textwidth]{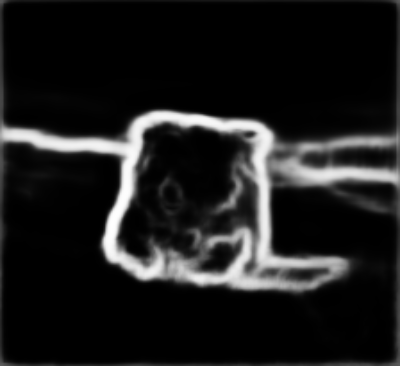}
		\includegraphics[width=0.17\textwidth]{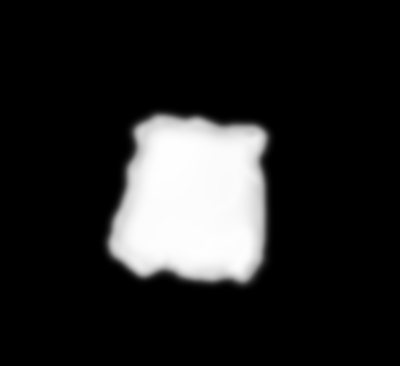}
		\includegraphics[width=0.17\textwidth]{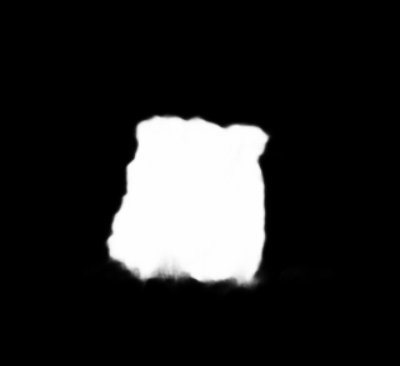}
		\\
		\begin{flushleft}
		    \vskip-12pt
			{\footnotesize \hskip45pt (a) \hskip50pt (b) \hskip50pt (c) \hskip50pt (d) \hskip50pt (e)}\\ 
		\end{flushleft}
		\vskip-5pt
		\includegraphics[width=0.17\textwidth]{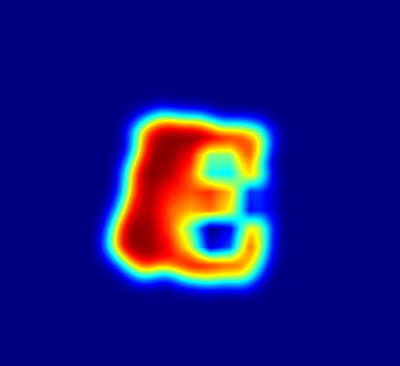}
		\includegraphics[width=0.17\textwidth]{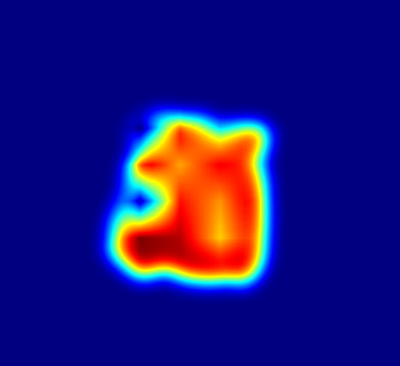}
		\includegraphics[width=0.17\textwidth]{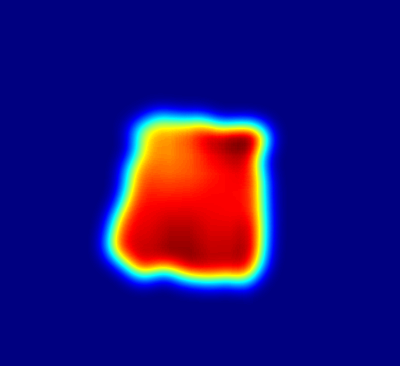}
		\includegraphics[width=0.17\textwidth]{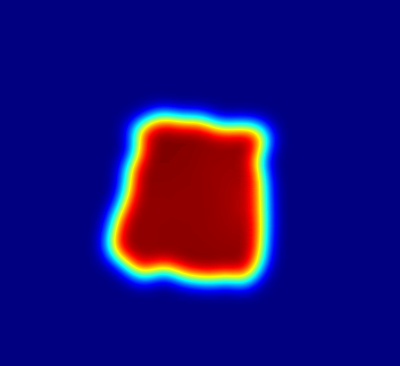}
		\includegraphics[width=0.17\textwidth]{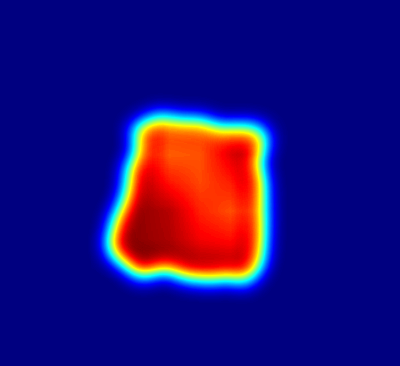}
		\\
		\begin{flushleft}
		    \vskip-12pt
			{\footnotesize \hskip45pt (f) \hskip50pt (g) \hskip50pt (h) \hskip50pt (i) \hskip50pt (j)}\\ 
		\end{flushleft}
		\vskip-15pt
		\caption{The CAM map comparisons with different patch-wise auxiliary classification tasks: (a) input image, (b) ground-truth, (c) edge map of the input image, (d) pseudo-GT computed using our 3SD method, (e) saliency map computed using 3SD where we use $32\times32$ pixel-wise classification, (f), (g), (h), (i), (j) are CAM-maps computed when auxiliary classification task is global class label, patchwise classification with patch size $16\times16$, $24\times24$, $32\times32$, $48\times48$ respectively.}
		\label{Fig:fig2_cmaps}
		\vspace{-2em}
\end{figure}

For the decoder, we follow the literature~\cite{caron2021emerging} by adding an  auxiliary classification task to capture important saliency cues like semantics and segmentation of the salient object in the image, which can be extracted in the form of class activation map (CAM map). However, performing a self-supervised classification with single global class label might result in low-quality class activation map (illustrated in Fig.~\ref{Fig:fig2_cmaps}f, where the CAM map is incomplete). To address this issue, we propose a contrastive learning~\cite{chen2020simple} based patch-wise self-supervision for the classification task, where we perform patch-wise ($32\times 32$ pixels in our implementation) classification and train it with proposed self-supervised contrastive loss. Specifically, positive patches (patches similar to the salient object) and negative patches (patches dissimilar to the salient object) are identified. Positive patches are pulled together, and they are pushed away from the negative patches. In this way, the network strives to learn the attentions or semantic information that are responsible for classifying the salient object at a fine-grain patch level.

With the novel designs of encoder and decoder, our 3SD is able to generate high-quality CAM maps (see Fig.~\ref{Fig:fig2_cmaps}). While the generated CAM maps provide salient object information, they might not have proper boundary corresponding to the salient object. To deal with this issue, we fuse CAM map with a gated edge map of the input image to generate the pseudo-GT salient map (Fig.~\ref{Fig:fig2_cmaps}d).


Fig.~\ref{Fig:fig1} compares sample results of the proposed 3SD with the existing SOTA weakly, unsupervised \cite{zhang2017supervision,zhang2018deep,zeng2019multi,zhang2020learning,zhang2020weakly} and SOTA fully-supervised \cite{CVPR2020_LDF} methods. One can clearly observe that \cite{zhang2017supervision,zhang2018deep,zeng2019multi,zhang2020learning,zhang2020weakly} fail to produce sharp edges and proper saliency maps. In contrast, our method is able to provide sharper and better results. To summarize, the main contributions of our paper are as follows:

\begin{itemize}
\item We propose a self-supervised 3SD method that requires no human annotations for training an SOD model. Our 3SD method is trained using high-quality pseudo-GT saliency maps generated from CAM maps using a novel self-supervised classification task.
\item We present a patch-wise self-supervised contrastive learning paradigm, which substantially improves the quality of pseudo-GT saliency maps and boosts 3SD performance.
\item We construct a novel encoder architecture for SOD that attends features locally (pixel-level understanding), and globally (patch-level understanding).
\item Extensive experiments on six benchmark datasets show that the proposed 3SD method outperforms the SOTA weakly/unsupervised methods.
\end{itemize}

\section{Related Work}
Classical image processing methods address SOD using histograms \cite{lu2013robust}, boundary connectivity \cite{zhu2014saliency}, high-dimensional color transforms \cite{kim2014salient}, hand-crafted features like foreground consistency \cite{zhang2017novel}, and similarity in superpixels \cite{zhang2019salient}.  In recent years, various supervised CNN-based methods have been proposed for SOD  \cite{zhang2017learning,luo2017non,wang2018detect,wu2019mutual,wu2019cascaded,feng2019attentive,Qin_2019_CVPR,CVPR2020_LDF} which extensively study architectural changes, attention mechanisms, multi-scale contextual information extraction, boundary-aware designs, label decoupling, \emph{etc}. In contrast, our 3SD method is a self-supervised method trained using pseudo-GT data. In what follows, we will review recent weakly/unsupervised SOD methods as well as the self-supervised methods.



\noindent\textbf{Weakly supervised SOD methods:} To reduce human efforts and expenses in pixel-level labeling and annotations, various weakly methods have been proposed. These methods use high-level labels such as image class labels and image captions ~\cite{li2018weakly,zeng2019multi}, and  scribble annotations \cite{zhang2020weakly}. Dai \etal\cite{dai2015boxsup}, Khorea \etal\cite{khoreva2017simple} follow a bounding-box label approach to solve weakly-supervised segmentation task. Wang \etal\cite{wang2017learning} extract cues using image-level labels for foreground salient objects. Hsu \etal\cite{hsu122017weakly} propose a category-based saliency map generator using image-level labels. \cite{chen2017deeplab,li2018weakly,obukhov2019gated} propose a CRF-based method for weakly supervised SOD.  Zeng \etal\cite{zeng2019multi} train a network with multiple source labels like category labels, and captions of the images to perform saliency detection. Zhang \etal\cite{zhang2020weakly} introduce scribble annotations for SOD. Unlike these methods, the proposed 3SD method doesn't require any kind of human annotations, noisy labels, scribble annotations, or hand-crafted features to perform SOD. 

 \noindent\textbf{Unsupervised SOD methods:}  Zhang \etal\cite{zhang2017supervision} devise a fusion process that employs unsupervised saliency models to generate supervision. Nguyen~\etal\cite{nguyen2019deepusps} propose an incrementally refinement technique that employs noisy pseudo labels generated from different handcrafted methods. Zhang \etal\cite{zhang2018deep} and Zhang \etal\cite{zhang2020learning} propose a saliency prediction network and noise modeling module that jointly learn from the noisy labels generated from multiple ``weak'' and ``noisy'' unsupervised handcrafted saliency methods. Unlike these unsupervised methods that highly rely on the noisy annotations, we propose a novel pseudo-GT generation technique using patchwise contrastive learning based self-supervised classification task.

\noindent\textbf{Self-supervised/Contrastive learning methods:} Several approaches explore discriminative approaches for self-supervised instance classification \cite{dosovitskiy2015discriminative,wu2018unsupervised}. These methods treat each image as a different class. The main limitation of these approaches is that they require comparing features from images for discrimination which can be complex in a large-scale dataset.   To address this challenge, Grill\etal \cite{grill2020bootstrap} introduce metric learning, called BOYL, where better representations of the features are learned by matching the outputs of momentum encoders. Subsequently, Caron \etal\cite{caron2021emerging} propose the DINO method that is based on mean Teacher \cite{tarvainen2017mean} self-distillation without labels. Recently, \cite{he2020momentum,chen2020simple,henaff2020data,tian2020contrastive} propose self-supervised contrastive learning based methods where the losses are inspired by noise-contrastive estimation\cite{gutmann2010noise}, triplet loss\cite{hermans2017defense}, and N-pair loss\cite{sohn2016improved}. 
Motivated by \cite{brendel2019approximating,chen2020simple}, we perform patch-wise contrastive learning within the self-supervised classification framework to obtain high-quality CAM maps, leading to good-quality pseudo-GTs.


\begin{figure}[h!]
    \centering
	\includegraphics[width=1.0\linewidth]{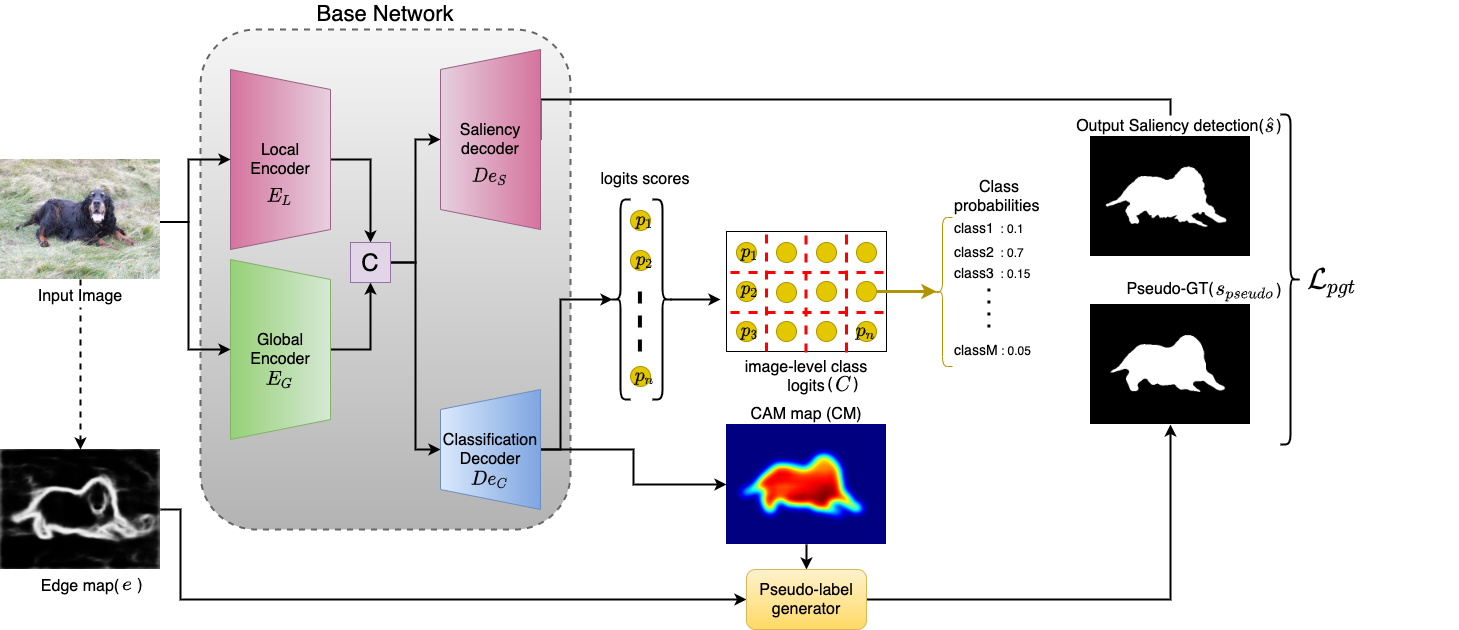} 
	\vskip -5pt 
	\caption{Overview of the 3SD method in computing the pseudo-GT for a given input image.}
	\label{fig:overview}
\end{figure}

\section{Proposed Method}
The proposed 3SD method is a fully self-supervised approach that doesn't require any human annotations or noisy labels. As shown in Fig.~\ref{fig:overview}, our method 3SD consists of a base network (BN) and a pseudo label generator. The base network outputs the saliency map along with the class labels which are used to generate the CAM map. BN utilizes self-supervised classification task to extract the semantic information for CAM map. The pseudo label generator fuses gated edge of the input image with CAM map to compute the pseudo-GT for training. In this section, we will discuss: (i) construction of the base network (BN), and (ii) pseudo-label generator. 

\subsection{Base Network}
We construct our framework with two encoders (local encoder $E_L$ and global encoder $E_G$) and two decoders (saliency decoder $De_{S}$ and classification decoder $De_{C}$) as shown in  Fig.~\ref{fig:overview}. Both local features and global context are vital for SOD task. To learn the pixel-wise relationship with local features, the local encoder $E_L$ is constructed using similar structure as U2Net \cite{qin2020u2}. Specifically, $E_L$ has nested two-level U-structure network with ResUBlock to capture contextual information at different scales. Even-though $E_L$ learns the inter-dependency between neighboring pixels in the receptive field size (approximately $96\times96$), it fails to capture the global context for high-resolution images. To remedy this issue, we introduce a transformer based encoder $E_G$ to model long-range relationship across patches (inspired by ViT\cite{dosovitskiy2020image}). By combining the outputs of both encoders, our 3SD is able to capture the local fine-grain details and reason globally. This combined encoded features are fed to saliency decoder $De_{S}$ to obtain saliency map, and classification decoder $De_{C}$ to obtain class label map as output. We use the similar architecture proposed in \cite{qin2020u2} for saliency decoder $De_{S}$. Our major contribution is on the classification decoder $De_{C}$. In contrast to conventional single-image-single-label classification design, our decoder performs self-supervised learning in patch wise. To enhance the feature representation and fully exploit the semantics, patches belonging to the object are encouraged to be differentiated from background patches using our novel contrastive learning paradigm.  
More details about $E_L$, $E_G$, $De_{S}$, and $De_{C}$ are provided in the supplementary document. 

As shown in Fig.~\ref{fig:overview}, given an input image, BN predicts the salient object $\hat{s}$, and patch-wise class logits map $C$. It is trained in a fully self-supervised way using our proposed pseudo-label generator, which is described in the next subsection.

\begin{figure}[ht!]
		\centering
		\includegraphics[width=0.16\textwidth]{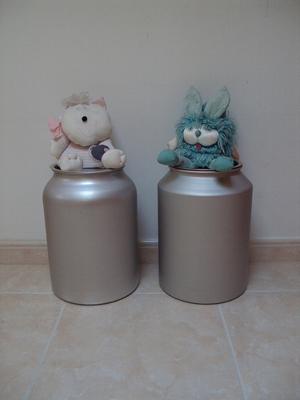}
		\includegraphics[width=0.16\textwidth]{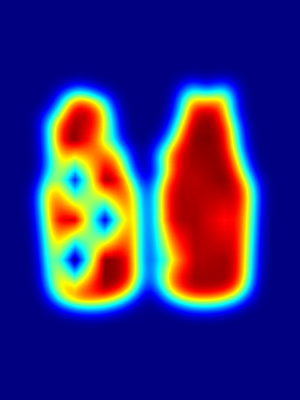}
		\includegraphics[width=0.16\textwidth]{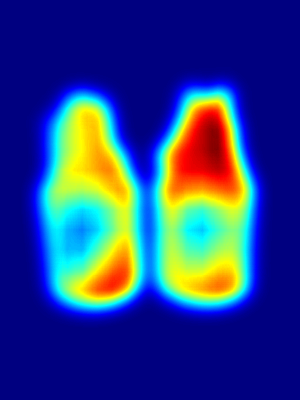}
		\includegraphics[width=0.16\textwidth]{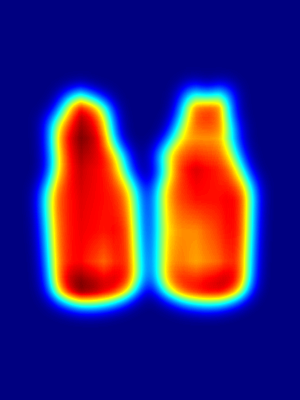}
		\includegraphics[width=0.16\textwidth]{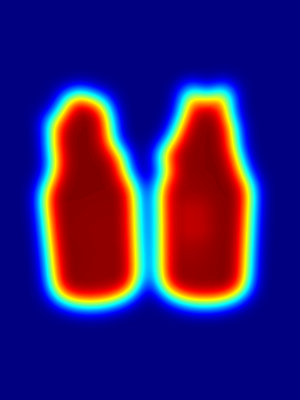}
		\includegraphics[width=0.16\textwidth]{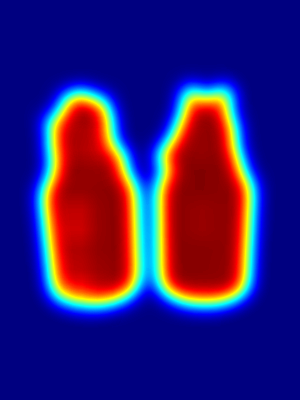}
		\\
		\includegraphics[width=0.16\textwidth]{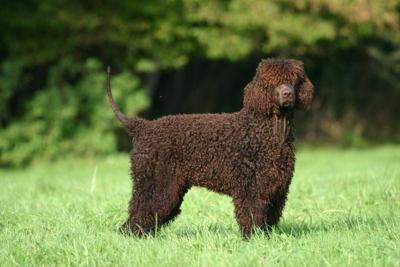}
		\includegraphics[width=0.16\textwidth]{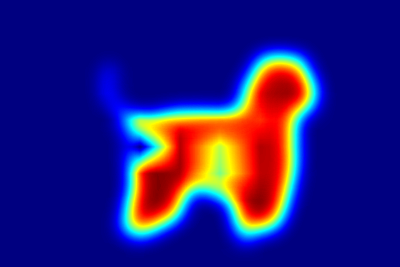}
		\includegraphics[width=0.16\textwidth]{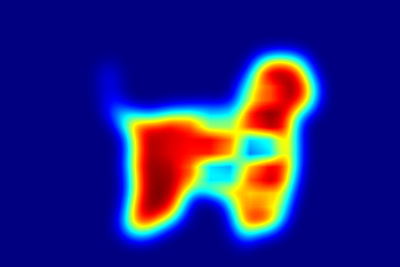}
		\includegraphics[width=0.16\textwidth]{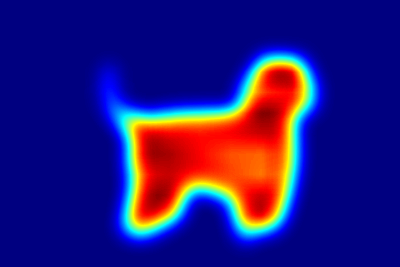}
		\includegraphics[width=0.16\textwidth]{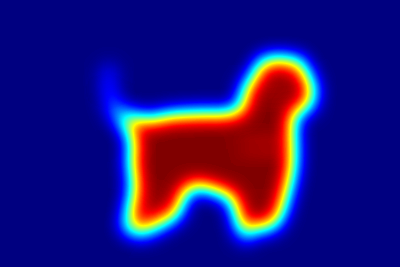}
		\includegraphics[width=0.16\textwidth]{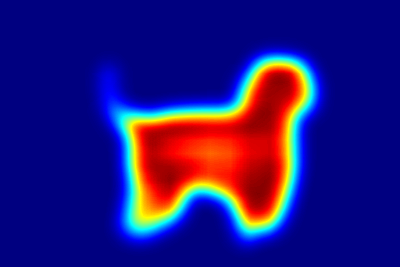}
		\\
		\includegraphics[width=0.16\textwidth]{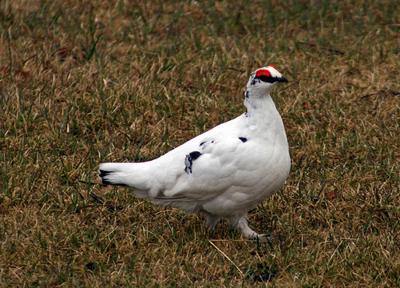}
		\includegraphics[width=0.16\textwidth]{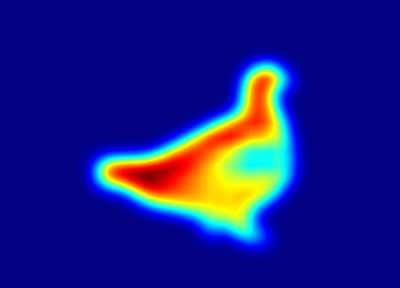}
		\includegraphics[width=0.16\textwidth]{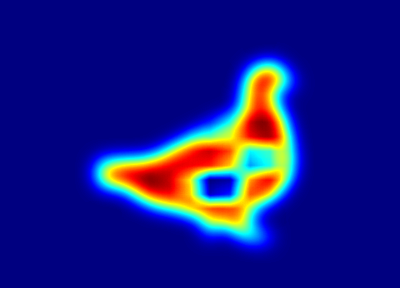}
		\includegraphics[width=0.16\textwidth]{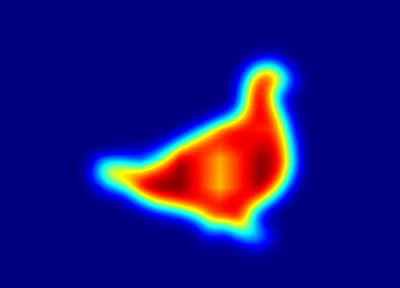}
		\includegraphics[width=0.16\textwidth]{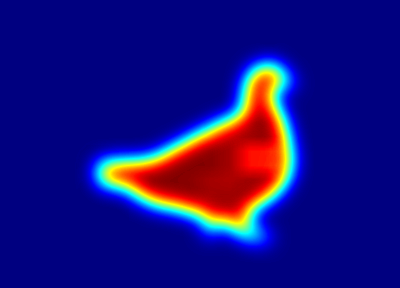}
		\includegraphics[width=0.16\textwidth]{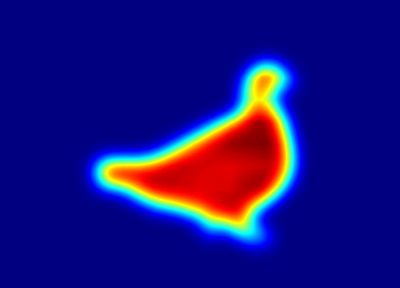}
		\\
		\begin{flushleft}
		    \vskip-12pt
			{\footnotesize \hskip25pt (a) \hskip45pt (b) \hskip45pt (c) \hskip46pt (d) \hskip46pt (e) \hskip46pt (f)}\\ 
		\end{flushleft}
		\vskip-15pt
		\caption{The CAM map comparison with different patch-wise auxiliary classification task. (a) input image,  (b) CAM-map computed when auxiliary classification is one global class label, patch-wise classification with patch size (c) $16\times16$, (d) $24\times24$, (e) $32\times32$, (f) $48\times48$.}
		\label{Fig:fig_cmaps}
\end{figure}

\subsection{Pseudo-label generator}\label{sec:pseudo_gt}
The main goal of 3SD is to train an SOD network without any GT-labels of salient objects. To achieve this, 3SD should be able to extract structural or semantic information of the object in the image. 
From the earlier works~\cite{Piao_2021_ICCV,li2018weakly} it is evident that attention maps from classification task provide important cues for salient object detection. In contrast to~\cite{Piao_2021_ICCV,li2018weakly} which require image class and/or caption labels,  
we train our BN with student-teacher based knowledge-distillation technique, and found the features from encoders of BN contain structural or semantic information of the salient object in the image. These semantics are the key to generate high-quality CAM maps \cite{zhou2016learning}. But as shown in Fig.~\ref{Fig:fig_cmaps}, the quality of CAM maps obtained from the self-supervised image-wise classification task when trained with single global class label, might not be high enough to produce pseudo-GT. This is due to the fact that single label classification task does not need to attend to all object regions. Instead, classification task drives the model to focus on the discriminative object parts. To address this issue, we propose a contrastive learning based patch-wise classification on the image patches as shown in Fig.~\ref{fig:selfdis_classification}. Patch-wise learning drives 3SD to capture local structures that constitute the object. Fig.~\ref{Fig:fig_cmaps} (b)-(e) show CAM map comparison between different classification tasks.
Finally, by guiding the CAM map with the edge information of the input image, we obtain high-quality pseudo-GT ($s_{pseudo}$). This pseudo-GT is taken as the training labels to update the parameters of the 3SD network.

\subsubsection{Self-supervised classification.} Student-teacher based knowledge distillation is a well-known learning paradigm that is commonly used for self-supervised classification tasks. As shown in Fig.~\ref{Fig:fig_cmaps}, self-supervised classification task as done in~\cite{caron2021emerging} fails to produce high quality pseudo-GTs. This is because image-wise self-supervised classification with one global class only requires a few important activations in the salient object, leading to incomplete saliency map. To overcome this hurdle, our proposed self-supervised classification contains (i) global level self-supervision using $\mathcal{L}_{st}$ loss, and (ii) patch-wise contrastive learning based self-supervision using $\mathcal{L}_{\rho}$ loss.


\begin{figure}[htp!]
    \centering
	\includegraphics[width=0.95\linewidth]{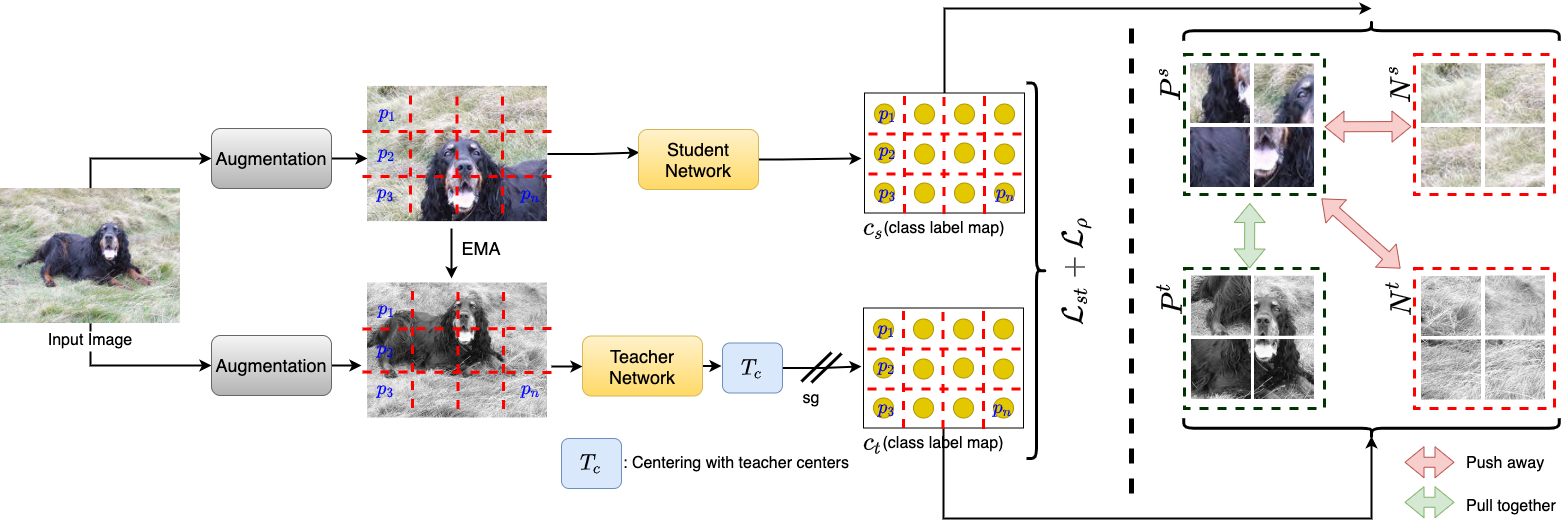} 
	\vskip -10pt 
	\caption{Self-distillation based training for classification. Here, ``ema" means EMA update rule used to update the teacher network's parameters, and ``sg" means stop gradient.}
	\label{fig:selfdis_classification}
\end{figure}

In our 3SD, student and teacher networks share the same architecture as the BN, and we denote student and teacher networks as $f_{\theta_s}$ and $f_{\theta_t}$, respectively with  $\theta_s$ and $\theta_t$ as their corresponding parameters. Given an input image, we compute class logits maps $C_s$ and $C_t$ for both networks, as well as their corresponding softmax probability output $P(C_s)$ and ${P(C_t)}$.
Meanwhile, we obtain image wise class logits for student model: $c_s = \sum_{\{p_i\}}C_s(p_i)$, 
where $p_i$ is $i^{th}$ patch in the image. Similar expression holds for the teacher's logits $c_t$. Given a fixed teacher network $f_{\theta_t}$, we match the image-level class probability distribution $P(c_s)$ and $P(c_t)$ by minimizing the cross-entropy loss to update the parameters of the student network $f_{\theta_s}$:
\begin{equation}
    \mathcal{L}_{st}~=~Q(P(c_s),P(c_t)),
\end{equation}
where $Q(a,b)=-a\log b$. Note that $\mathcal{L}_{st}$ is used to update the student network's parameters $\theta_s$ with stochastic gradient descent. The teacher weights are updated using exponential moving average (EMA) update rule as in \cite{caron2021emerging,grill2020bootstrap}, \textit{i.e.}, 
$\theta_t \leftarrow\lambda\theta_t + (1-\lambda)\theta_s$, with $\lambda$ following a cosine schedule from 0.996 to 1 during training.
Additionally, in order to improve quality of CAM maps we perform patchwise contrastive learning for the classification task, where we identify the top $M_{\rho}$ positive patches (patches similar to salient object) and $M_{\rho}$ negative patches (patches dissimilar to salient object) from $C_s$ and $C_t$. Using these positive and negative pairs we construct the following contrastive learning~\cite{chen2020simple} based loss,
\begin{equation}
\mathcal{L}_{\rho}=\frac{1}{M_{\rho}^2}\sum_{\{p^s_i\} \in P^s }\sum_{\{p^t_j\} \in P^t}-\log \frac{\exp \left(\operatorname{sim}\left(C_s(p^s_i), C_t(p^t_j)\right) / \tau\right)}{\sum_{n_k \in N^t,N^s}^{}  \exp \left(\operatorname{sim}\left(C_s(p^s_i), C_t(n_k)\right) / \tau\right)}
\end{equation}
where $P^s$ and $N^s$ are the set of $M_{\rho}$ positive and negative patches from $C_s$ respectively, and $P^t$ and $N^t$ are the set of $M_{\rho}$ positive and negative patches from $C_t$ respectively. Concretely, $P^s$ and $N^s$ are extracted from $C_s$ by comparing $C_s(p_i)$ with $c_s$ using normalized dot product. Similarly, $P^t$ and $N^t$ are obtained from $C_t$. From Fig.~\ref{Fig:fig_cmaps}, we can observe that addition of loss $\mathcal{L}_{\rho}$ drives the student network to extract class-sensitive local features across the whole object region, which results in better quality CAM maps.

In summary, the following steps are performed in self-supervised classification task to learn student network weights: (i) Augment the image with two different augmentations and randomly crop the images to obtain $x'^l$ (local crop or smaller resolution) and $x''^g$ (global crop or smaller resolution) as shown in  Fig.~\ref{fig:selfdis_classification}. (ii) Fix the teacher network's parameters. (iii) Pass $x'^l$ as input to the student network and $x''^g$ as input to the teacher network, to obtain output class logits maps $C'_s$ and $C''_t$, respectively. (iv) Match the outputs by minimizing the loss $\mathcal{L}_{st}$ and update the student network's parameters $\theta_s$. (v) Additionally, we extract positive and negative patches from $C_s$ and $C_t$, and update the student network's parameters $\theta_s$ using $\mathcal{L}_{\rho}$. (v) Finally, employ the EMA update rule to update the teacher network's parameters $\theta_t$. Fig.~\ref{fig:selfdis_classification} shows the overview diagram of these steps in updating the student network's parameters. 

\subsubsection{Pseudo-GT}
\begin{figure}[h!]
		\centering
		\includegraphics[width=0.24\textwidth]{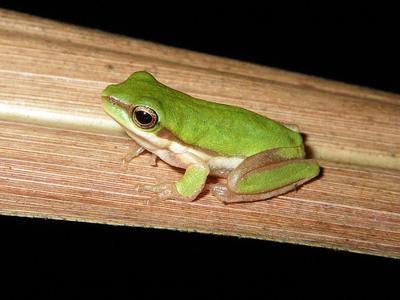}
		\includegraphics[width=0.24\textwidth]{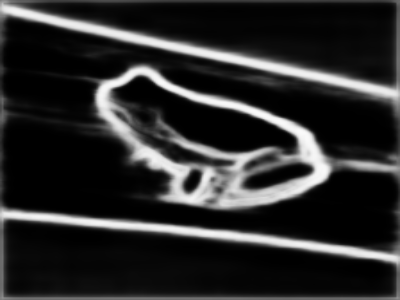}
		\includegraphics[width=0.24\textwidth]{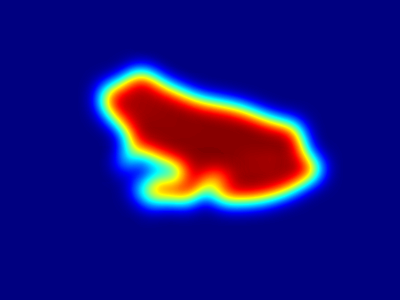}
		\includegraphics[width=0.24\textwidth]{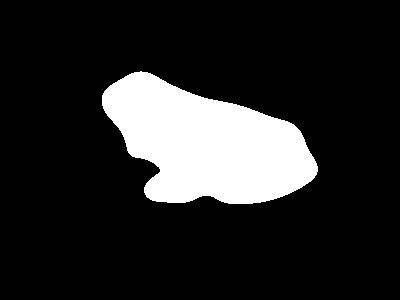}
		\\
		\begin{flushleft}
		    \vskip-12pt
			{\footnotesize \hskip40pt (a) \hskip72pt (b) \hskip72pt (c) \hskip72pt (d)}\\ 
		\end{flushleft}
		\vskip-5pt
		\includegraphics[width=0.24\textwidth]{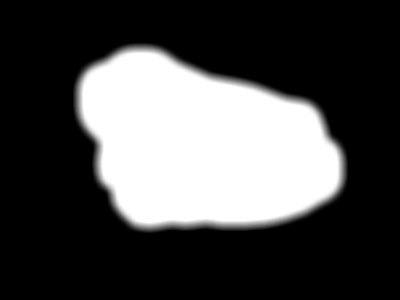}
		\includegraphics[width=0.24\textwidth]{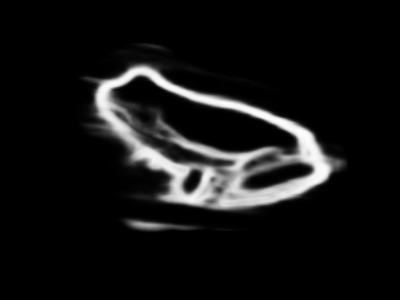}
		\includegraphics[width=0.24\textwidth]{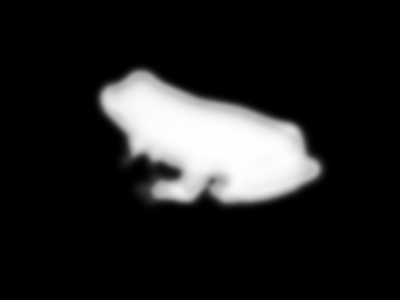}
		\includegraphics[width=0.24\textwidth]{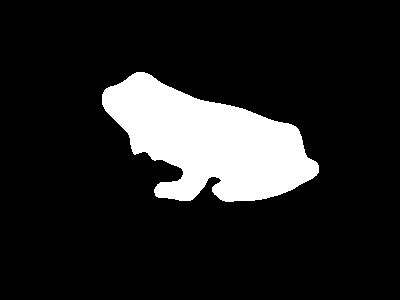}
		\\
		\begin{flushleft}
		    \vskip-12pt
			{\footnotesize \hskip40pt (e) \hskip72pt (f) \hskip72pt (g) \hskip72pt (h)}\\ 
		\end{flushleft}
		\vskip-20pt
		\caption{(a) input image, (b) edge map (c) CAM-maps computed by 3SD, (d) thresholded cam map, (e) dialated cam map, (f) gated edge map, (g) pseudo-GT, (h) thresholded pseudo-GT.}
		\label{Fig:pseudo_GT}
\end{figure}

When 3SD is trained with self-supervised classification, features from classification decoder explicitly contain the structural or semantic information of the salient object in the image. Hence, we compute CAM (using technique from \cite{zhou2016learning}) to obtain this semantic information of the salient object. We can clearly observe in Fig.~\ref{Fig:pseudo_GT} (c) and (d), that the obtained CAM map doesn't depict a proper or sharp boundary. To overcome this problem, we compute edge map using the RCF edge detection network \cite{liu2017richer}. Then we dilate the thresholded CAM map and gate the edge map to obtain the edges of the salient objects. This design effectively removes the background edges. Pseudo-GT ($s_{pseudo}$) is then defined as the union of the gated edge map and the CAM map, \textit{i.e.}, $s_{pseudo} = CM \cup g_e$, where $CM$ is the CAM map and $g_e$ is the gated edge. Fig.~\ref{Fig:pseudo_GT} (g) and (h) show the pseudo-GT ($s_{pseudo}$) and the thresholded version of the pseudo-GT ($s_{pseudo}$) (with threshold 0.5). We compute cross-entropy between the pseudo-GT ($s_{pseudo}$) and the salient object $\hat{s}$ prediction output of 3SD, 
\begin{equation}
    \resizebox{0.91\hsize}{!}{ $\mathcal{L}_{pgt} = -\sum_{(j,k)}^{(H,W)} \left[s_{pseudo}(j,k)\log\hat{s}(j,k)\right. \left. +\left(1-s_{pseudo}(j,k)\right)\log\left(1-\hat{s}(j,k))\right)\right],$}
\end{equation}
where $(j,k)$ are pixel coordinates, and $(H,W)$ are height and width of the image.
We minimize $\mathcal{L}_{p}$ and update the 3SD parameters to perform SOD.

\subsection{Loss}
To improve the boundary estimation for SOD, we further introduce the gated structure-aware loss ($\mathcal{L}_{gs}$) proposed by \cite{zhang2020weakly}. Gated structure-aware ($\mathcal{L}_{gs}$) is defined as follows,
\begin{equation}
    \mathcal{L}_{gs} = \sum_{(i,j)}\sum_{d\in(\overrightarrow{h},\overrightarrow{v})} \Psi\left(|\partial_d s(i,j)|exp\left(-0.5|\partial_d(g.x(i,j))|\right) \right),
\end{equation}
where $\Psi(s) = \sqrt{s^2+e^{-6}}$, $(i,j)$ are pixel coordinates, $(\overrightarrow{h},\overrightarrow{v})$ are horizontal and vertical directions, $\partial$ is derivative operation, and $g$ is the gate for the structure loss (see Fig.~\ref{Fig:pseudo_GT}(e)). 

The overall loss function used for training the 3SD network is defined as follows,
\begin{equation}
    \mathcal{L}_{total} = \mathcal{L}_{st} + \mathcal{L}_{\rho}+ \mathcal{L}_{pgt} + \beta_1 \mathcal{L}_{gs}(\hat{s},x).
\end{equation}
We set $\beta_1=0.3$ in all our experiments.

\section{Experiments and results}
In this section, we present the details of various experiments conducted on six benchmarks to demonstrate the effectiveness of the proposed 3SD method. Additionally, we perform extensive ablation studies which shows the factors and parameters that influence the performance of the proposed 3SD method.
\subsection{Datasets}
\noindent\textbf{Training dataset:}
We use images from the DUTS-TR dataset which is the training dataset of DUTS~\cite{wang2017learning}. Note that we only use the images to train 3SD, \textit{i.e.}, we don't use the corresponding salient object annotations. The DUTS-TR dataset contains 10,553 images, and it is the largest and most commonly used dataset for training SOD.

\noindent\textbf{Evaluation datasets:} We evaluate the performance of  3SD on six benchmark test datasets: (i) The DUTS-TE  testing dataset (5,019 images) \cite{wang2017learning}, (ii) DUT-OMRON  (5,168 test images) \cite{yang2013saliency}, (iii) HKU-IS \cite{li2015visual} (4,447 test images), (iv) PASCAL-S \cite{li2014secrets} (850 test images), (v) ECSSD \cite{yan2013hierarchical} (1,000 images), and (vi) THUR \cite{cheng2014salientshape} (6,233 images). 
\vspace{-1.0em}
\subsection{Comparison methods and metrics}
\vspace{-0.5em}
We compare the performance of 3SD with seven SOTA weakly/unsupervised methods \cite{zhang2017supervision,li2018weakly,zhang2018deep,zeng2019multi,zhang2020weakly,zhang2020learning}, and eleven SOTA fully-supervised methods \cite{zhang2017learning,luo2017non,wang2018detect,liu2018picanet,zhang2018progressive,wu2019mutual,wu2019cascaded,feng2019attentive,Qin_2019_CVPR,qin2020u2,CVPR2020_LDF}. We use the following five evaluation metrics for comparison: structure measure~\cite{fan2017structure} ($S_m$), saliency structure measure~\cite{zhang2020weakly} ($B_\mu$), mean F-score ($F_\beta$), mean E-measure\cite{fan2018enhanced} ($E_\eta$), and mean absolute error ($MAE$). Additionally, we plot precision vs. recall curves to show the effectiveness of our 3SD method.
\subsection{Implementation Details}
\vspace{-0.5em}
Given a training image, we randomly crop $320\times 320$ pixels and use it for training  3SD. As explained in the earlier sections, we perform patch-wise contrastive learning based self-supervised classification  without labels. Specifically, we learn the teacher centers $T_c$ ($K$ dimension vector) to sharpen the teacher's output before computing $\mathcal{L}_{st}$. Here, we sharpen the teachers output class logits map with $T_c$ as shown in Fig~\ref{fig:selfdis_classification} (~\ie~ $C_t(p_i) = C_t(p_i)+T_c$). In all of our experiments, we perform $32\times32$ patch-wise self-supervised classification in our 3SD method. For augmentations, we use the augmentations of BYOL~\cite{grill2020bootstrap} (Gaussian blur
solarization, and color jittering) and multi-crop \cite{caron2020unsupervised} with
bi-cubic interpolation to resize the crop. In our 3SD method, the student and teacher networks share the same architecture as BN with different weights.
Initially, we train the student and teacher networks for self-supervised classification as explained in section~\ref{sec:pseudo_gt} for 30 epochs. Later, we perform patch-wise self-supervised classification and pseudo-GT based salient object detection in an iterative manner as explained in the algorithm provided in the supplementary document. We set $M_{\rho}$ as 10 for all our experiments.  
We train 3SD for 150 epochs using a stochastic gradient descent optimizer with batch-size 8 and learning rate 0.001. Following DUTS~\cite{wang2017learning}, we set $K=200$ classes in all of our experiments. The student network ($f_{\theta_s}$) in 3SD is used for all evaluations during inference. Algorithm~\ref{alg:train} shows steps involved in training our 3SD method.

\begin{algorithm}[h!]
		\SetAlgoLined
		\KwIn{Set of training images $\mathcal{D}=\{x^i\}_{i=1}^N$, $T_c$ teacher centers $K$ dimension vector. }
		\KwResult{$\hat{\theta}_{s}$, optimized network parameters of 3SD}
		\For{every epoch }{
			\For{$\{x\}\in\mathcal{D}$}{
				\#\#\# self-supervised classification \#\#\#\\
				\# augmentation images \#\\
				$x',\:x^{\prime \prime}\: =\: $ augment1($x$), augment2($x$)\\
				$x'^l$,$x^{\prime \prime g}$ = localview($x'$), globalview($x^{\prime \prime}$)\\
				$s_{s}, C_{s}=f_{\theta_{s}}\left(x^{\prime l}\right)$ \# student output\\
                $s_{t}, C_{t}=f_{\theta_{t}}\left(x^{\prime \prime g}\right)$ \# teacher output\\
                extract $P^s$ and $N^s$ from $C_s$; extract $P^t$ and $N^t$ from $C_t$\\
                compute $\mathcal{L}_{s c}+\mathcal{L}_{\rho} ;$ update $\theta_{s}$\\
                \# EMA update teacher weights and centers \#\\
                $\theta_{t} \leftarrow \lambda \theta_{t}+(1-\lambda) \theta_{s} ; \lambda \in[0.996,1]$\\
                $T_c \leftarrow \lambda T_c+(1-\lambda) T_{c,t}$\\
                \#\#\# Pseudo-GT generation \#\#\#\\
                compute CAM map ($CM$) for image $x$ using classification $f_{\theta_{s}}$\\
                obtain $g_e$ using $CM$ and edge $e$ (using \cite{liu2017richer}) \\
                $s_{p s e u d o}=C M \cup g_{e}$\\
                \#\#\# update student network weights $f_{\theta_{s}}$ for SOD task \#\#\#\\
                $\hat{s}, \hat{c}=f_{\theta_{s}}(x)$\\
                using $s_{pseudo}$ compute $\mathcal{L}_{pgt}$
                compute $\mathcal{L}_{b}(\hat{s}, x)$\\
                update $\theta_{s}$ using $\mathcal{L}_{pgt}+\mathcal{L}_{b}(\hat{s}, x)$
			}
		}
		\caption{Pseudo code for training 3SD.}\label{alg:train}
\end{algorithm}

\subsection{Comparisons}
\noindent\textbf{Quantitative results:} As shown in Table~\ref{tab:comp_sota}, our 3SD method consistently outperforms the SOTA weakly/unsupervised methods on all datasets and in all metrics. This shows training with our pseudo-GT generation is superior to the existing weakly/unsupervised SOD techniques. Moreover, although 3SD is not trained with any human annotations or weak labels (\emph{e.g.} image captions, handcrafted features, or scribble annotations), its performance is on par with fully-supervised methods. In some cases the performance of 3SD is even better than some fully-supervised methods like ~\cite{luo2017non,liu2018picanet,zhang2018progressive}. Similar behaviors can be observed in the  precision vs. recall curves shown in Fig.~\ref{fig:precisionvsrecall}. Curves corresponding to our method are close to the SOTA fully-supervised methods, while the curves corresponding to the SOTA weakly/unsupervised methods are far below our 3SD method.

\begin{figure*}[h!]
		\centering
		\includegraphics[width=0.094\textwidth]{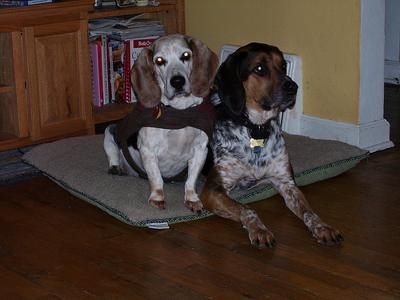}
		\includegraphics[width=0.094\textwidth]{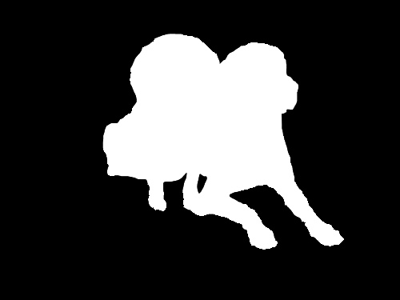}
		\includegraphics[width=0.094\textwidth]{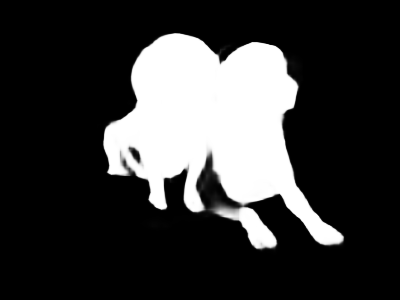}
		\includegraphics[width=0.094\textwidth]{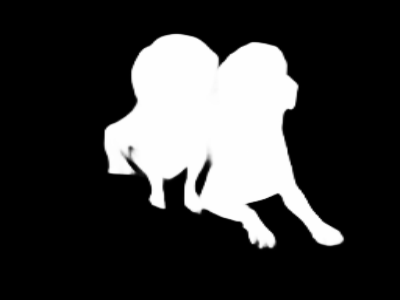}
		\includegraphics[width=0.094\textwidth]{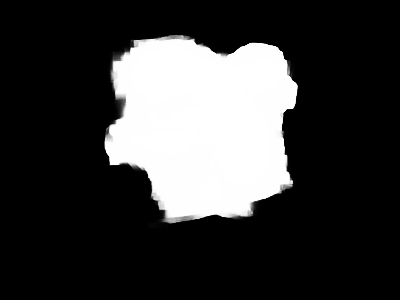}
		\includegraphics[width=0.094\textwidth]{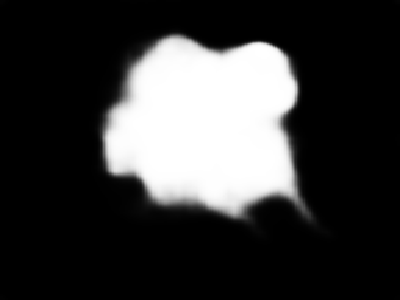}
		\includegraphics[width=0.094\textwidth]{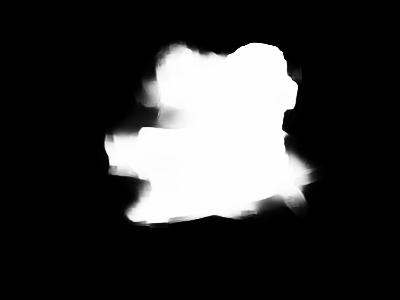}
		\includegraphics[width=0.094\textwidth]{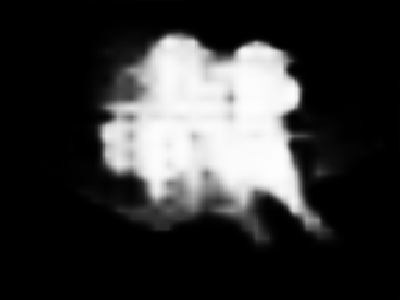}
		\includegraphics[width=0.094\textwidth]{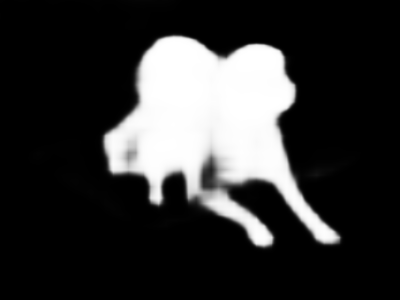}
		\includegraphics[width=0.094\textwidth]{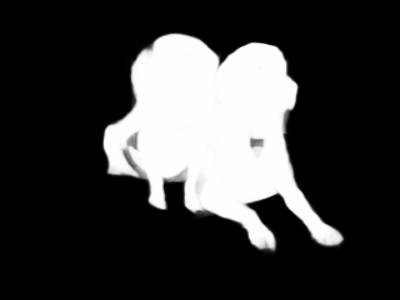}
		\\
		\includegraphics[width=0.094\textwidth]{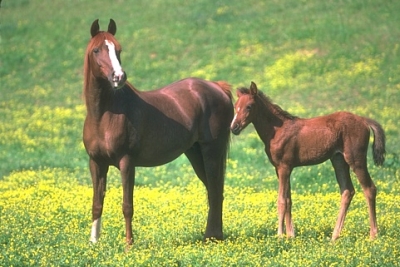}
		\includegraphics[width=0.094\textwidth]{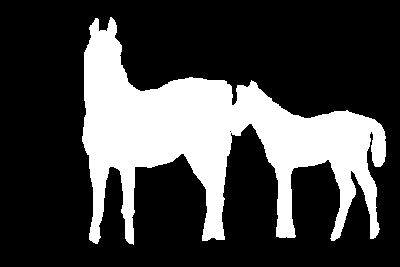}
		\includegraphics[width=0.094\textwidth]{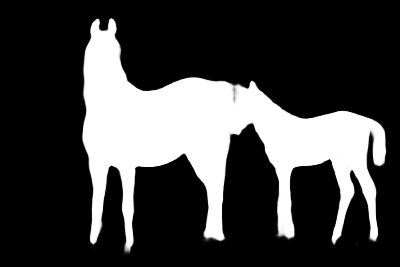}
		\includegraphics[width=0.094\textwidth]{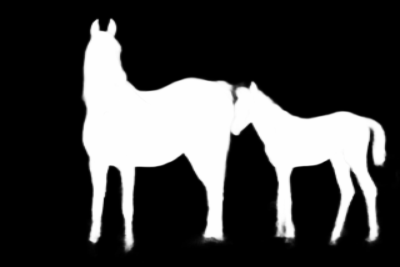}
		\includegraphics[width=0.094\textwidth]{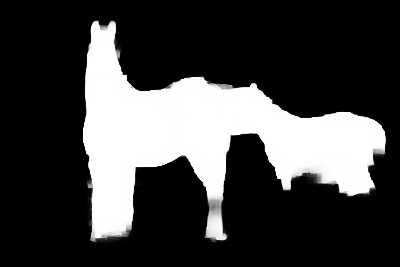}
		\includegraphics[width=0.094\textwidth]{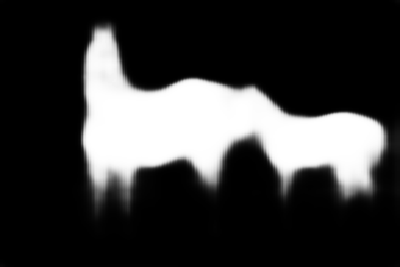}
		\includegraphics[width=0.094\textwidth]{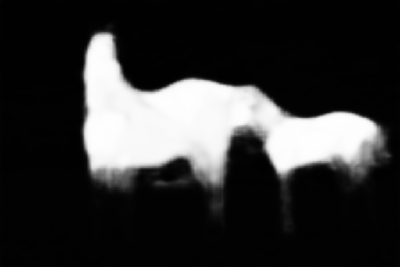}
		\includegraphics[width=0.094\textwidth]{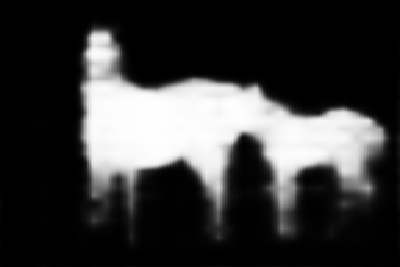}
		\includegraphics[width=0.094\textwidth]{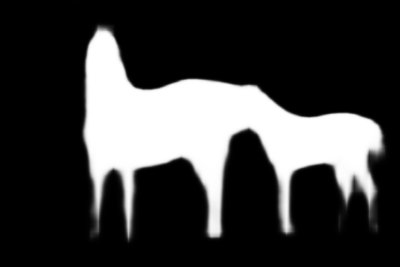}
		\includegraphics[width=0.094\textwidth]{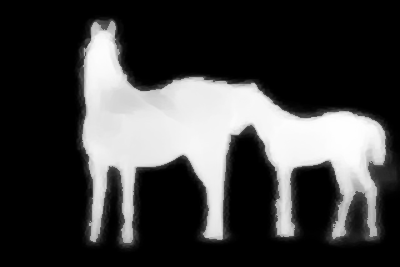}
		\\
		
		\includegraphics[width=0.094\textwidth]{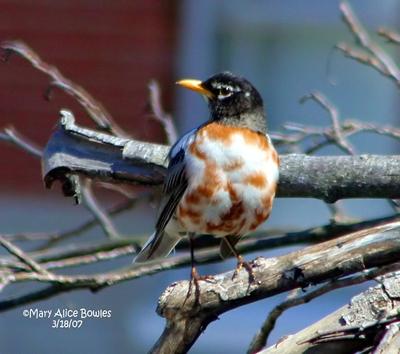}
		\includegraphics[width=0.094\textwidth]{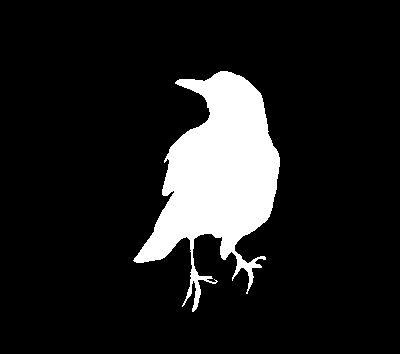}
        \includegraphics[width=0.094\textwidth]{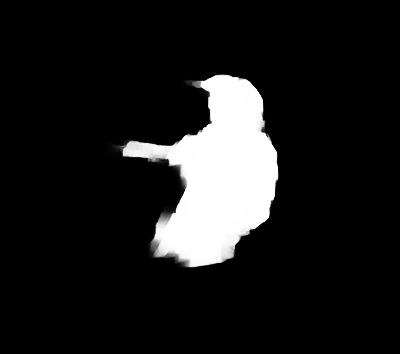}
        \includegraphics[width=0.094\textwidth]{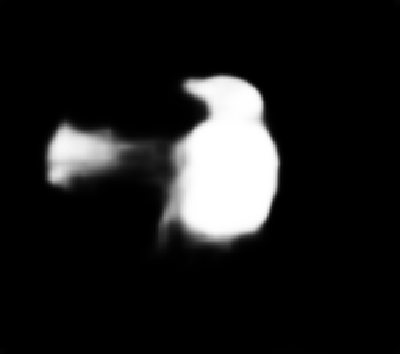}
		\includegraphics[width=0.094\textwidth]{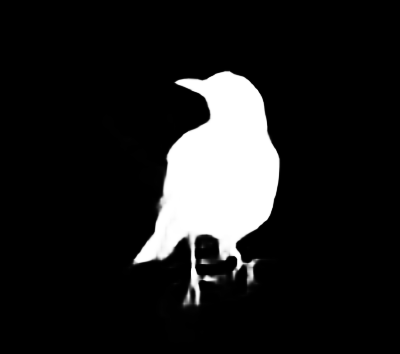}
		\includegraphics[width=0.094\textwidth]{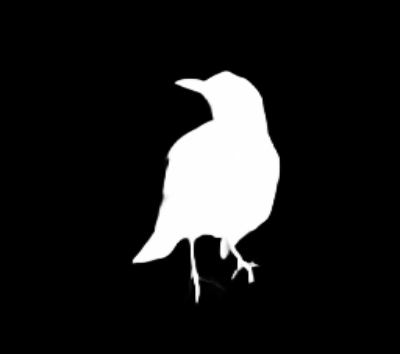}
		\includegraphics[width=0.094\textwidth]{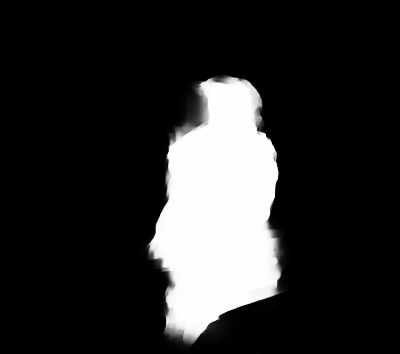}
		\includegraphics[width=0.094\textwidth]{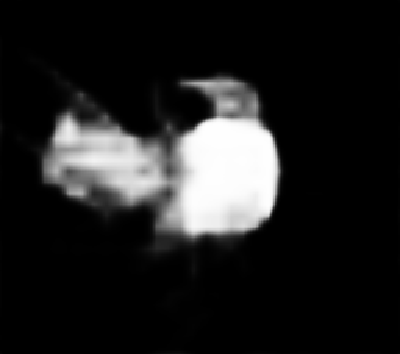}
		\includegraphics[width=0.094\textwidth]{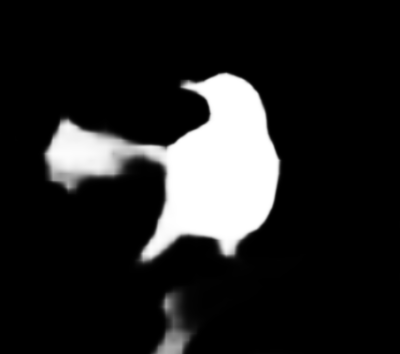}
		\includegraphics[width=0.094\textwidth]{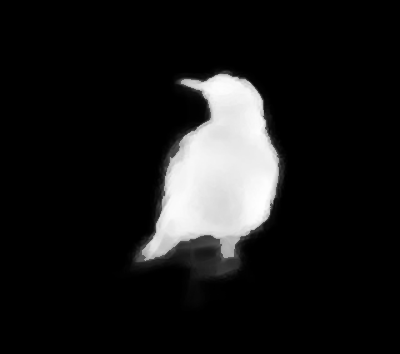}
		\\
		\includegraphics[width=0.094\textwidth]{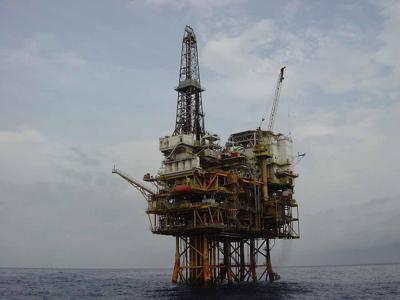}
		\includegraphics[width=0.094\textwidth]{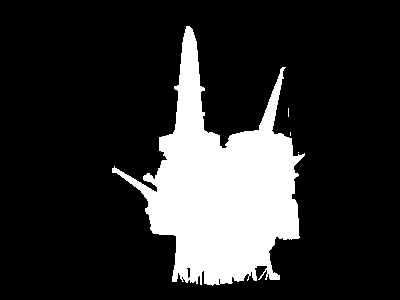}
		\includegraphics[width=0.094\textwidth]{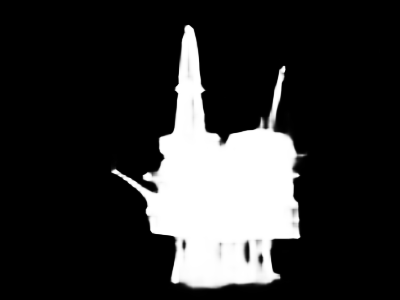}
		\includegraphics[width=0.094\textwidth]{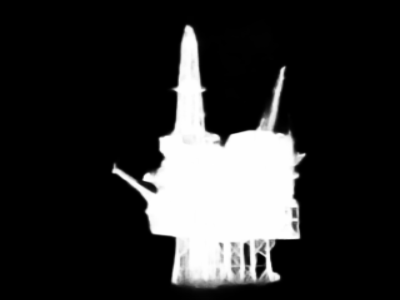}
		\includegraphics[width=0.094\textwidth]{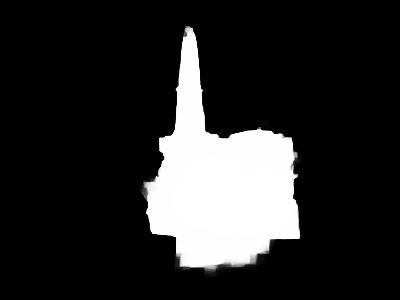}
		\includegraphics[width=0.094\textwidth]{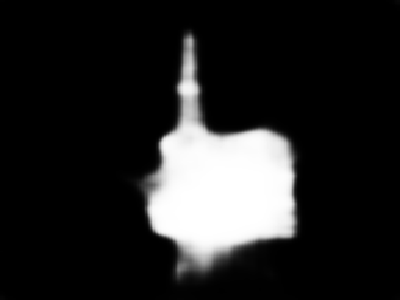}
		\includegraphics[width=0.094\textwidth]{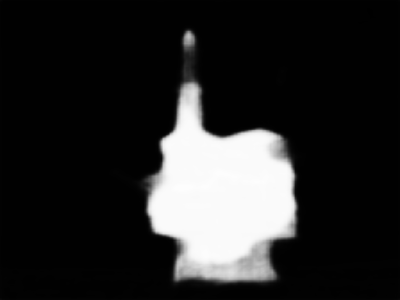}
		\includegraphics[width=0.094\textwidth]{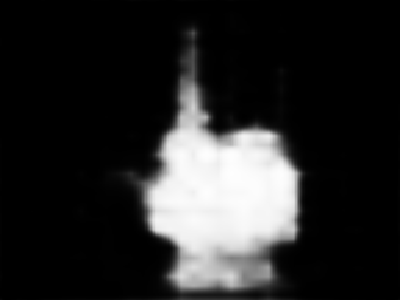}
		\includegraphics[width=0.094\textwidth]{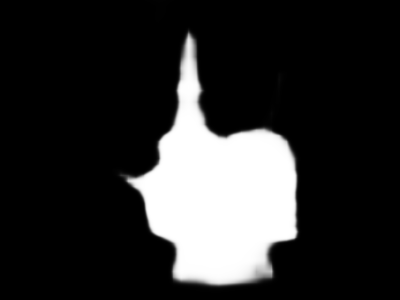}
		\includegraphics[width=0.094\textwidth]{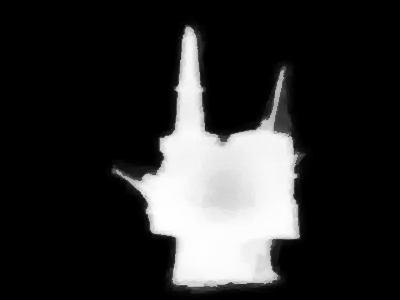}
		\\
		\begin{flushleft}
		    \vskip-12pt
			{\footnotesize \hskip5pt Image \hskip17pt GT \hskip20pt \cite{CVPR2020_LDF}\hskip7pt BN(ours) \hskip10pt \cite{zhang2020weakly} \hskip20pt \cite{Piao_2021_ICCV}  \hskip15pt \cite{zhang2017supervision} \hskip17pt \cite{zeng2019multi}\hskip20pt \cite{zhang2020learning}\hskip5pt 3SD(ours)}\\ 
		\end{flushleft}
		\vskip-20pt
		\caption{Qualitative comparison of 3SD method against SOTA fully-supervised, and SOTA weakly/unsupervised methods.}
		\label{Fig:comparison_sota}
\end{figure*}

\begin{figure*}[h!]
	\begin{center}
		\includegraphics[width=\linewidth]{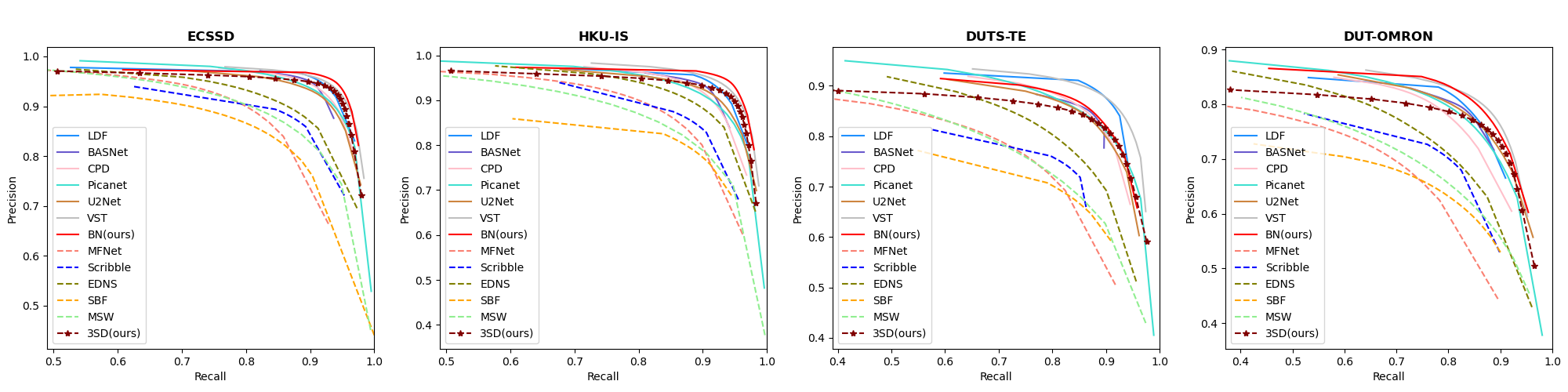} 
	\end{center}
	\vskip -20pt 
	\caption{Precision vs. recall curve comparison of our method with SOTA methods on the ECSSD, HKU-IS, PASCAL, DUTS-TE, and DUT-OMRON  datasets. In the precision vs. recall graphs, we represent all supervised methods with thick lines and weakly/unsupervised methods with dotted lines.}
	\label{fig:precisionvsrecall}
\end{figure*}

\noindent\textbf{Qualitative results:} Fig.~\ref{Fig:comparison_sota} illustrates the qualitative comparisons of our 3SD method with SOTA methods on 4 sample images from DUTS-TE, DUT-OMRON, HKU-IS, and ECSSD. It can be seen that the outputs of \cite{zhang2017supervision,zeng2019multi,zhang2020weakly,zhang2020learning} are blurred or incomplete, and include parts of non-salient objects. In contrast, 3SD outputs are accurate, clear, and sharp. For example,  output saliency maps of \cite{zhang2017supervision,zeng2019multi,zhang2020weakly,zhang2020learning} in the second row of Fig.~\ref{Fig:comparison_sota} miss parts of legs for the horses. And, output saliency maps of \cite{zhang2017supervision,zeng2019multi,zhang2020weakly,zhang2020learning} in the third row of Fig.~\ref{Fig:comparison_sota} contains artifacts or parts of non-salient objects. In contrast, saliency maps of our 3SD method delineate legs for the horses properly, and are free of artifacts. More qualitative comparisons are provided in the supplementary material.

\begin{table}[htp!]
\caption{Comparison of our method with SOTA methods on six benchmark datasets (DUTS-TE, DUT-OMRON, HKU-IS, PASCAL, ECSSD and THUR) using the metrics  $S_m,\: B_{\mu},\: F_{\beta},\: E_{\eta},$ and $MAE$ where $\uparrow \& \downarrow$ denote larger and smaller is better, respectively.  }
\vskip-10pt
\label{tab:comp_sota}
\resizebox{1\linewidth}{!}{
\begin{tabular}{|l|c|cccccccc|ccccc|cccc|}
\hline{\color[HTML]{000000} } & {\color[HTML]{000000} } & \multicolumn{8}{c|}{{\color[HTML]{000000} Fully supervised}} & \multicolumn{5}{c|}{{\color[HTML]{000000} Weakly supervised}} & \multicolumn{4}{c|}{{\color[HTML]{000000} Un-supervised}} \\ \cline{3-19} 
\multirow{-2}{*}{{\color[HTML]{000000} Dataset}} & \multirow{-2}{*}{{\color[HTML]{000000} Metric}} & {\color[HTML]{000000} \begin{tabular}[c]{@{}c@{}}PiCANet\cite{liu2018picanet}\\ CVPR18\end{tabular}} & {\color[HTML]{000000} \begin{tabular}[c]{@{}c@{}}MSNet\cite{wu2019mutual}\\ CVPR19\end{tabular}} & {\color[HTML]{000000} \begin{tabular}[c]{@{}c@{}}CPD\cite{wu2019cascaded}\\ CVPR19\end{tabular}} & {\color[HTML]{000000} \begin{tabular}[c]{@{}c@{}}BASNet\cite{Qin_2019_CVPR}\\ CVPR19\end{tabular}} & {\color[HTML]{000000} \begin{tabular}[c]{@{}c@{}}U2Net\cite{qin2020u2}\\ PR2020\end{tabular}} & {\color[HTML]{000000} \begin{tabular}[c]{@{}c@{}}LDF\cite{CVPR2020_LDF}\\ CVPR20\end{tabular}} & \begin{tabular}[c]{@{}c@{}}VST\cite{liu2021visual}\\ ICCV21\end{tabular} & {\color[HTML]{000000} \textbf{\begin{tabular}[c]{@{}c@{}}BN\\ (ours)\end{tabular}}} & {\color[HTML]{000000} \begin{tabular}[c]{@{}c@{}}WSS\cite{wang2017learning}\\ CVPR17\end{tabular}} & {\color[HTML]{000000} \begin{tabular}[c]{@{}c@{}}WSI\cite{li2018weakly}\\ AAAI18\end{tabular}} & \begin{tabular}[c]{@{}c@{}}MSW\cite{zeng2019multi}\\ CVPR19\end{tabular} & {\color[HTML]{000000} \begin{tabular}[c]{@{}c@{}}Scrible\_S\cite{zhang2020weakly}\\ CVPR20\end{tabular}} & {\color[HTML]{000000} \begin{tabular}[c]{@{}c@{}}MFNet\cite{Piao_2021_ICCV}\\ ICCV21\end{tabular}} & {\color[HTML]{000000} \begin{tabular}[c]{@{}c@{}}SBF\cite{zhang2017supervision}\\ ICCV17\end{tabular}} & {\color[HTML]{000000} \begin{tabular}[c]{@{}c@{}}MNL\cite{zhang2018deep}\\ CVPR18\end{tabular}} & {\color[HTML]{000000} \begin{tabular}[c]{@{}c@{}}EDNS\cite{zhang2020learning}\\ ECCV20\end{tabular}} & {\color[HTML]{000000} \textbf{\begin{tabular}[c]{@{}c@{}}3SD\\ (ours)\end{tabular}}} \\ \hline
{\color[HTML]{000000} } & {\color[HTML]{000000} $S_m\uparrow$} & {\color[HTML]{000000} 0.851} & {\color[HTML]{000000} 0.851} & {\color[HTML]{000000} 0.867} & {\color[HTML]{000000} 0.866} & {\color[HTML]{000000} 0.861} & {\color[HTML]{000000} 0.881} & 0.885 & {\color[HTML]{000000} 0.883} & {\color[HTML]{000000} 0.748} & {\color[HTML]{000000} 0.697} & 0.759 & {\color[HTML]{000000} 0.793} & {\color[HTML]{000000} 0.775} & {\color[HTML]{000000} 0.739} & {\color[HTML]{000000} 0.813} & {\color[HTML]{000000} 0.828} & {\color[HTML]{000000} {\ul \textbf{0.844}}} \\
{\color[HTML]{000000} } & {\color[HTML]{000000} $B_{\mu}\downarrow$} & {\color[HTML]{000000} 0.635} & {\color[HTML]{000000} 0.582} & {\color[HTML]{000000} 0.462} & {\color[HTML]{000000} 0.400} & {\color[HTML]{000000} 0.365} & {\color[HTML]{000000} 0.384} & 0.347 & {\color[HTML]{000000} 0.358} & {\color[HTML]{000000} 0.780} & {\color[HTML]{000000} 0.879} & 0.829 & {\color[HTML]{000000} 0.603} & {\color[HTML]{000000} 0.610} & {\color[HTML]{000000} 0.808} & {\color[HTML]{000000} 0.712} & {\color[HTML]{000000} 0.628} & {\color[HTML]{000000} {\ul \textbf{0.443}}} \\
{\color[HTML]{000000} } & {\color[HTML]{000000} $F_{\beta}\uparrow$} & {\color[HTML]{000000} 0.757} & {\color[HTML]{000000} 0.792} & {\color[HTML]{000000} 0.8246} & {\color[HTML]{000000} 0.823} & {\color[HTML]{000000} 0.804} & {\color[HTML]{000000} 0.855} & 0.870 & {\color[HTML]{000000} 0.829} & {\color[HTML]{000000} 0.633} & {\color[HTML]{000000} 0.569} & 0.648 & {\color[HTML]{000000} 0.746} & {\color[HTML]{000000} 0.770} & {\color[HTML]{000000} 0.622} & {\color[HTML]{000000} 0.725} & {\color[HTML]{000000} 0.747} & {\color[HTML]{000000} {\ul \textbf{0.765}}} \\
{\color[HTML]{000000} } & {\color[HTML]{000000} $E_{\eta}\uparrow$} & {\color[HTML]{000000} 0.853} & {\color[HTML]{000000} 0.883} & {\color[HTML]{000000} 0.902} & {\color[HTML]{000000} 0.896} & {\color[HTML]{000000} 0.897} & {\color[HTML]{000000} 0.910} & 0.939 & {\color[HTML]{000000} 0.913} & {\color[HTML]{000000} 0.806} & {\color[HTML]{000000} 0.690} & 0.742 & {\color[HTML]{000000} 0.865} & {\color[HTML]{000000} 0.839} & {\color[HTML]{000000} 0.763} & {\color[HTML]{000000} 0.853} & {\color[HTML]{000000} 0.859} & {\color[HTML]{000000} {\ul \textbf{0.875}}} \\
\multirow{-5}{*}{{\color[HTML]{000000} DUTS-TE}} & {\color[HTML]{000000} $MAE\downarrow$} & {\color[HTML]{000000} 0.062} & {\color[HTML]{000000} 0.050} & {\color[HTML]{000000} 0.043} & {\color[HTML]{000000} 0.048} & {\color[HTML]{000000} 0.044} & {\color[HTML]{000000} 0.034} & 0.037 & {\color[HTML]{000000} 0.036} & {\color[HTML]{000000} 0.100} & {\color[HTML]{000000} 0.116} & 0.091 & {\color[HTML]{000000} 0.062} & {\color[HTML]{000000} 0.076} & {\color[HTML]{000000} 0.107} & {\color[HTML]{000000} 0.075} & {\color[HTML]{000000} 0.060} & {\color[HTML]{000000} {\ul \textbf{0.044}}} \\ \hline
{\color[HTML]{000000} } & {\color[HTML]{000000} $S_m\uparrow$} & {\color[HTML]{000000} 0.826} & {\color[HTML]{000000} 0.809} & {\color[HTML]{000000} 0.818} & {\color[HTML]{000000} 0.836} & {\color[HTML]{000000} 0.842} & {\color[HTML]{000000} 0.847} & 0.839 & {\color[HTML]{000000} 0.843} & {\color[HTML]{000000} 0.730} & {\color[HTML]{000000} 0.759} & 0.756 & {\color[HTML]{000000} 0.771} & {\color[HTML]{000000} 0.742} & {\color[HTML]{000000} 0.731} & {\color[HTML]{000000} 0.733} & {\color[HTML]{000000} 0.791} & {\color[HTML]{000000} {\ul \textbf{0.803}}} \\
{\color[HTML]{000000} } & {\color[HTML]{000000} $B_{\mu}\downarrow$} & {\color[HTML]{000000} 0.685} & {\color[HTML]{000000} 0.642} & {\color[HTML]{000000} 0.549} & {\color[HTML]{000000} 0.480} & {\color[HTML]{000000} 0.438} & {\color[HTML]{000000} 0.432} & 0.455 & {\color[HTML]{000000} 0.429} & {\color[HTML]{000000} 0.830} & {\color[HTML]{000000} 0.839} & 0.890 & {\color[HTML]{000000} 0.655} & {\color[HTML]{000000} 0.658} & {\color[HTML]{000000} 0.812} & {\color[HTML]{000000} 0.776} & {\color[HTML]{000000} 0.689} & {\color[HTML]{000000} {\ul \textbf{0.501}}} \\
{\color[HTML]{000000} } & {\color[HTML]{000000} $F_{\beta}\uparrow$} & {\color[HTML]{000000} 0.710} & {\color[HTML]{000000} 0.709} & {\color[HTML]{000000} 0.739} & {\color[HTML]{000000} 0.767} & {\color[HTML]{000000} 0.757} & {\color[HTML]{000000} 0.773} & 0.800 & {\color[HTML]{000000} 0.777} & {\color[HTML]{000000} 0.590} & {\color[HTML]{000000} 0.641} & 0.597 & {\color[HTML]{000000} 0.702} & {\color[HTML]{000000} 0.646} & {\color[HTML]{000000} 0.612} & {\color[HTML]{000000} 0.597} & {\color[HTML]{000000} 0.701} & {\color[HTML]{000000} {\ul \textbf{0.735}}} \\
{\color[HTML]{000000} } & {\color[HTML]{000000} $E_{\eta}\uparrow$} & {\color[HTML]{000000} 0.823} & {\color[HTML]{000000} 0.830} & {\color[HTML]{000000} 0.845} & {\color[HTML]{000000} 0.865} & {\color[HTML]{000000} 0.867} & {\color[HTML]{000000} 0.873} & 0.883 & {\color[HTML]{000000} 0.869} & {\color[HTML]{000000} 0.729} & {\color[HTML]{000000} 0.761} & 0.728 & {\color[HTML]{000000} 0.835} & {\color[HTML]{000000} 0.803} & {\color[HTML]{000000} 0.763} & {\color[HTML]{000000} 0.712} & {\color[HTML]{000000} 0.816} & {\color[HTML]{000000} {\ul \textbf{0.845}}} \\
\multirow{-5}{*}{{\color[HTML]{000000} DUT-OMRON}} & {\color[HTML]{000000} $MAE\downarrow$} & {\color[HTML]{000000} 0.072} & {\color[HTML]{000000} 0.064} & {\color[HTML]{000000} 0.057} & {\color[HTML]{000000} 0.057} & {\color[HTML]{000000} 0.054} & {\color[HTML]{000000} 0.051} & 0.058 & {\color[HTML]{000000} 0.049} & {\color[HTML]{000000} 0.110} & {\color[HTML]{000000} 0.100} & 0.109 & {\color[HTML]{000000} 0.068} & {\color[HTML]{000000} 0.087} & {\color[HTML]{000000} 0.108} & {\color[HTML]{000000} 0.103} & {\color[HTML]{000000} 0.070} & {\color[HTML]{000000} {\ul \textbf{0.660}}} \\ \hline
{\color[HTML]{000000} } & {\color[HTML]{000000} $S_m\uparrow$} & {\color[HTML]{000000} 0.906} & {\color[HTML]{000000} 0.907} & {\color[HTML]{000000} 0.904} & {\color[HTML]{000000} 0.909} & {\color[HTML]{000000} 0.916} & {\color[HTML]{000000} 0.919} & 0.919 & {\color[HTML]{000000} 0.918} & {\color[HTML]{000000} 0.822} & {\color[HTML]{000000} 0.808} & 0.818 & {\color[HTML]{000000} 0.855} & {\color[HTML]{000000} 0.846} & {\color[HTML]{000000} 0.812} & {\color[HTML]{000000} 0.860} & {\color[HTML]{000000} 0.890} & {\color[HTML]{000000} {\ul \textbf{0.904}}} \\
{\color[HTML]{000000} } & {\color[HTML]{000000} $B_{\mu}\downarrow$} & {\color[HTML]{000000} 0.561} & {\color[HTML]{000000} 0.498} & {\color[HTML]{000000} 0.421} & {\color[HTML]{000000} 0.359} & {\color[HTML]{000000} 0.325} & {\color[HTML]{000000} 0.337} & 0.338 & {\color[HTML]{000000} 0.321} & {\color[HTML]{000000} 0.752} & {\color[HTML]{000000} 0.782} & 0.830 & {\color[HTML]{000000} 0.537} & {\color[HTML]{000000} 0.522} & {\color[HTML]{000000} 0.734} & {\color[HTML]{000000} 0.627} & {\color[HTML]{000000} 0.567} & {\color[HTML]{000000} {\ul \textbf{0.381}}} \\
{\color[HTML]{000000} } & {\color[HTML]{000000} $F_{\beta}\uparrow$} & {\color[HTML]{000000} 0.854} & {\color[HTML]{000000} 0.878} & {\color[HTML]{000000} 0.895} & {\color[HTML]{000000} 0.903} & {\color[HTML]{000000} 0.890} & {\color[HTML]{000000} 0.914} & 0.922 & {\color[HTML]{000000} 0.918} & {\color[HTML]{000000} 0.773} & {\color[HTML]{000000} 0.763} & 0.734 & {\color[HTML]{000000} 0.857} & {\color[HTML]{000000} 0.851} & {\color[HTML]{000000} 0.783} & {\color[HTML]{000000} 0.820} & {\color[HTML]{000000} 0.878} & {\color[HTML]{000000} {\ul \textbf{0.885}}} \\
{\color[HTML]{000000} } & {\color[HTML]{000000} $E_{\eta}\uparrow$} & {\color[HTML]{000000} 0.909} & {\color[HTML]{000000} 0.930} & {\color[HTML]{000000} 0.940} & {\color[HTML]{000000} 0.943} & {\color[HTML]{000000} 0.945} & {\color[HTML]{000000} 0.954} & 0.962 & {\color[HTML]{000000} 0.959} & {\color[HTML]{000000} 0.819} & {\color[HTML]{000000} 0.800} & 0.786 & {\color[HTML]{000000} 0.923} & {\color[HTML]{000000} 0.921} & {\color[HTML]{000000} 0.855} & {\color[HTML]{000000} 0.858} & {\color[HTML]{000000} 0.919} & {\color[HTML]{000000} {\ul \textbf{0.935}}} \\
\multirow{-5}{*}{{\color[HTML]{000000} HKU-IS}} & {\color[HTML]{000000} $MAE\downarrow$} & {\color[HTML]{000000} 0.047} & {\color[HTML]{000000} 0.039} & {\color[HTML]{000000} 0.033} & {\color[HTML]{000000} 0.032} & {\color[HTML]{000000} 0.031} & {\color[HTML]{000000} 0.027} & 0.030 & {\color[HTML]{000000} 0.024} & {\color[HTML]{000000} 0.079} & {\color[HTML]{000000} 0.089} & 0.084 & {\color[HTML]{000000} 0.047} & {\color[HTML]{000000} 0.059} & {\color[HTML]{000000} 0.075} & {\color[HTML]{000000} 0.065} & {\color[HTML]{000000} 0.043} & {\color[HTML]{000000} {\ul \textbf{0.039}}} \\ \hline
{\color[HTML]{000000} } & {\color[HTML]{000000} $S_m\uparrow$} & {\color[HTML]{000000} 0.848} & {\color[HTML]{000000} 0.844} & {\color[HTML]{000000} 0.848} & {\color[HTML]{000000} 0.838} & {\color[HTML]{000000} 0.844} & {\color[HTML]{000000} 0.851} & 0.863 & {\color[HTML]{000000} 0.856} & {\color[HTML]{000000} -} & {\color[HTML]{000000} -} & 0.697 & {\color[HTML]{000000} 0.742} & {\color[HTML]{000000} 0.770} & {\color[HTML]{000000} 0.712} & {\color[HTML]{000000} 0.728} & {\color[HTML]{000000} 0.750} & {\color[HTML]{000000} {\ul \textbf{0.761}}} \\
{\color[HTML]{000000} } & {\color[HTML]{000000} $B_{\mu}\downarrow$} & {\color[HTML]{000000} 0.704} & {\color[HTML]{000000} 0.671} & {\color[HTML]{000000} 0.616} & {\color[HTML]{000000} 0.582} & {\color[HTML]{000000} 0.513} & {\color[HTML]{000000} 0.512} & 0.504 & {\color[HTML]{000000} 0.509} & {\color[HTML]{000000} 0.831} & {\color[HTML]{000000} 0.855} & 0.870 & {\color[HTML]{000000} 0.665} & {\color[HTML]{000000} 0.679} & {\color[HTML]{000000} 0.815} & {\color[HTML]{000000} 0.776} & {\color[HTML]{000000} 0.739} & {\color[HTML]{000000} {\ul \textbf{0.531}}} \\
{\color[HTML]{000000} } & {\color[HTML]{000000} $F_{\beta}\uparrow$} & {\color[HTML]{000000} 0.799} & {\color[HTML]{000000} 0.813} & {\color[HTML]{000000} 0.822} & {\color[HTML]{000000} 0.821} & {\color[HTML]{000000} 0.797} & {\color[HTML]{000000} 0.848} & 0.829 & {\color[HTML]{000000} 0.832} & {\color[HTML]{000000} 0.698} & {\color[HTML]{000000} 0.653} & 0.685 & {\color[HTML]{000000} 0.788} & {\color[HTML]{000000} 0.751} & {\color[HTML]{000000} 0.735} & {\color[HTML]{000000} 0.748} & {\color[HTML]{000000} 0.759} & {\color[HTML]{000000} {\ul \textbf{0.763}}} \\
{\color[HTML]{000000} } & {\color[HTML]{000000} $E_{\eta}\uparrow$} & {\color[HTML]{000000} 0.804} & {\color[HTML]{000000} 0.822} & {\color[HTML]{000000} 0.819} & {\color[HTML]{000000} 0.821} & {\color[HTML]{000000} 0.831} & {\color[HTML]{000000} 0.865} & 0.865 & {\color[HTML]{000000} 0.849} & {\color[HTML]{000000} 0.690} & {\color[HTML]{000000} 0.647} & 0.693 & {\color[HTML]{000000} 0.798} & {\color[HTML]{000000} 0.817} & {\color[HTML]{000000} 0.746} & {\color[HTML]{000000} 0.741} & {\color[HTML]{000000} 0.794} & {\color[HTML]{000000} {\ul \textbf{0.810}}} \\
\multirow{-5}{*}{{\color[HTML]{000000} PASCAL}} & {\color[HTML]{000000} $MAE\downarrow$} & {\color[HTML]{000000} 0.129} & {\color[HTML]{000000} 0.119} & {\color[HTML]{000000} 0.122} & {\color[HTML]{000000} 0.122} & {\color[HTML]{000000} 0.074} & {\color[HTML]{000000} 0.060} & 0.067 & {\color[HTML]{000000} 0.068} & {\color[HTML]{000000} 0.184} & {\color[HTML]{000000} 0.206} & 0.178 & {\color[HTML]{000000} 0.140} & {\color[HTML]{000000} 0.115} & {\color[HTML]{000000} 0.167} & {\color[HTML]{000000} 0.158} & {\color[HTML]{000000} 0.142} & {\color[HTML]{000000} {\ul \textbf{0.137}}} \\ \hline
{\color[HTML]{000000} } & {\color[HTML]{000000} $S_m\uparrow$} & {\color[HTML]{000000} 0.867} & {\color[HTML]{000000} 0.905} & {\color[HTML]{000000} 0.905} & {\color[HTML]{000000} 0.910} & {\color[HTML]{000000} 0.918} & {\color[HTML]{000000} 0.912} & 0.917 & {\color[HTML]{000000} 0.924} & {\color[HTML]{000000} 0.808} & {\color[HTML]{000000} 0.805} & 0.825 & {\color[HTML]{000000} 0.854} & {\color[HTML]{000000} 0.834} & {\color[HTML]{000000} 0.813} & {\color[HTML]{000000} 0.845} & {\color[HTML]{000000} 0.860} & {\color[HTML]{000000} \textbf{0.883}} \\
{\color[HTML]{000000} } & {\color[HTML]{000000} $B_{\mu}\downarrow$} & {\color[HTML]{000000} 0.592} & {\color[HTML]{000000} 0.542} & {\color[HTML]{000000} 0.434} & {\color[HTML]{000000} 0.364} & {\color[HTML]{000000} 0.321} & {\color[HTML]{000000} 0.323} & 0.348 & {\color[HTML]{000000} 0.319} & {\color[HTML]{000000} 0.808} & {\color[HTML]{000000} 0.801} & 0.851 & {\color[HTML]{000000} 0.550} & {\color[HTML]{000000} 0.581} & {\color[HTML]{000000} 0.759} & {\color[HTML]{000000} 0.681} & {\color[HTML]{000000} 0.600} & {\color[HTML]{000000} {\ul \textbf{0.387}}} \\
{\color[HTML]{000000} } & {\color[HTML]{000000} $F_{\beta}\uparrow$} & {\color[HTML]{000000} 0.871} & {\color[HTML]{000000} 0.885} & {\color[HTML]{000000} 0.907} & {\color[HTML]{000000} 0.913} & {\color[HTML]{000000} 0.910} & {\color[HTML]{000000} 0.930} & 0.929 & {\color[HTML]{000000} 0.933} & {\color[HTML]{000000} 0.767} & {\color[HTML]{000000} 0.762} & 0.761 & {\color[HTML]{000000} 0.865} & {\color[HTML]{000000} 0.854} & {\color[HTML]{000000} 0.782} & {\color[HTML]{000000} 0.810} & {\color[HTML]{000000} 0.852} & {\color[HTML]{000000} {\ul \textbf{0.887}}} \\
{\color[HTML]{000000} } & {\color[HTML]{000000} $E_{\eta}\uparrow$} & {\color[HTML]{000000} 0.909} & {\color[HTML]{000000} 0.922} & {\color[HTML]{000000} 0.932} & {\color[HTML]{000000} 0.938} & {\color[HTML]{000000} 0.936} & {\color[HTML]{000000} 0.925} & 0.945 & {\color[HTML]{000000} 0.940} & {\color[HTML]{000000} 0.796} & {\color[HTML]{000000} 0.792} & 0.787 & {\color[HTML]{000000} 0.908} & {\color[HTML]{000000} 0.885} & {\color[HTML]{000000} 0.835} & {\color[HTML]{000000} 0.836} & {\color[HTML]{000000} 0.883} & {\color[HTML]{000000} {\ul \textbf{0.928}}} \\
\multirow{-5}{*}{{\color[HTML]{000000} ECSSD}} & {\color[HTML]{000000} $MAE\downarrow$} & {\color[HTML]{000000} 0.054} & {\color[HTML]{000000} 0.048} & {\color[HTML]{000000} 0.043} & {\color[HTML]{000000} 0.040} & {\color[HTML]{000000} 0.033} & {\color[HTML]{000000} 0.034} & 0.034 & {\color[HTML]{000000} 0.032} & {\color[HTML]{000000} 0.108} & {\color[HTML]{000000} 0.068} & 0.098 & {\color[HTML]{000000} 0.061} & {\color[HTML]{000000} 0.084} & {\color[HTML]{000000} 0.096} & {\color[HTML]{000000} 0.090} & {\color[HTML]{000000} 0.071} & {\color[HTML]{000000} {\ul \textbf{0.049}}} \\ \hline
{\color[HTML]{000000} } & {\color[HTML]{000000} $S_m\uparrow$} & {\color[HTML]{000000} -} & {\color[HTML]{000000} 0.819} & {\color[HTML]{000000} 0.831} & {\color[HTML]{000000} 0.823} & {\color[HTML]{000000} 0.828} & {\color[HTML]{000000} 0.847} & - & {\color[HTML]{000000} 0.852} & {\color[HTML]{000000} 0.775} & {\color[HTML]{000000} -} & - & {\color[HTML]{000000} 0.794} & {\color[HTML]{000000} -} & {\color[HTML]{000000} 0.762} & {\color[HTML]{000000} 0.804} & {\color[HTML]{000000} 0.810} & {\color[HTML]{000000} {\ul \textbf{0.827}}} \\
{\color[HTML]{000000} } & {\color[HTML]{000000} $B_{\mu}\downarrow$} & {\color[HTML]{000000} 0.659} & {\color[HTML]{000000} 0.620} & {\color[HTML]{000000} 0.525} & {\color[HTML]{000000} 0.489} & {\color[HTML]{000000} 0.531} & {\color[HTML]{000000} 0.469} & - & {\color[HTML]{000000} 0.464} & {\color[HTML]{000000} 0.788} & {\color[HTML]{000000} -} & - & {\color[HTML]{000000} 0.596} & {\color[HTML]{000000} -} & {\color[HTML]{000000} 0.785} & {\color[HTML]{000000} 0.717} & {\color[HTML]{000000} 0.656} & {\color[HTML]{000000} {\ul \textbf{0.487}}} \\
{\color[HTML]{000000} } & {\color[HTML]{000000} $F_{\beta}\uparrow$} & {\color[HTML]{000000} 0.710} & {\color[HTML]{000000} 0.718} & {\color[HTML]{000000} 0.750} & {\color[HTML]{000000} 0.737} & {\color[HTML]{000000} 0.749} & {\color[HTML]{000000} 0.764} & - & {\color[HTML]{000000} 0.769} & {\color[HTML]{000000} 0.653} & {\color[HTML]{000000} -} & - & {\color[HTML]{000000} 0.718} & {\color[HTML]{000000} -} & {\color[HTML]{000000} 0.627} & {\color[HTML]{000000} 0.691} & {\color[HTML]{000000} 0.719} & {\color[HTML]{000000} {\ul \textbf{0.746}}} \\
{\color[HTML]{000000} } & {\color[HTML]{000000} $E_{\eta}\uparrow$} & {\color[HTML]{000000} 0.821} & {\color[HTML]{000000} 0.829} & {\color[HTML]{000000} 0.851} & {\color[HTML]{000000} 0.841} & {\color[HTML]{000000} 0.843} & {\color[HTML]{000000} 0.842} & - & {\color[HTML]{000000} 0.846} & {\color[HTML]{000000} 0.775} & {\color[HTML]{000000} -} & - & {\color[HTML]{000000} 0.837} & {\color[HTML]{000000} -} & {\color[HTML]{000000} 0.770} & {\color[HTML]{000000} 0.807} & {\color[HTML]{000000} 0.838} & {\color[HTML]{000000} {\ul \textbf{0.840}}} \\
\multirow{-5}{*}{{\color[HTML]{000000} THUR}} & {\color[HTML]{000000} $MAE\downarrow$} & {\color[HTML]{000000} 0.084} & {\color[HTML]{000000} 0.079} & {\color[HTML]{000000} 0.094} & {\color[HTML]{000000} 0.073} & {\color[HTML]{000000} 0.075} & {\color[HTML]{000000} 0.064} & - & {\color[HTML]{000000} 0.061} & {\color[HTML]{000000} 0.097} & {\color[HTML]{000000} -} & - & {\color[HTML]{000000} 0.077} & {\color[HTML]{000000} -} & {\color[HTML]{000000} 0.107} & {\color[HTML]{000000} 0.086} & {\color[HTML]{000000} 0.070} & {\color[HTML]{000000} {\ul \textbf{0.064}}} \\ \hline
{\color[HTML]{000000} Parameters} & {\color[HTML]{000000} Size in MB} & {\color[HTML]{000000} 197} & {\color[HTML]{000000} -} & {\color[HTML]{000000} 183} & {\color[HTML]{000000} 349} & {\color[HTML]{000000} 176} & {\color[HTML]{000000} 98} & 82 & {\color[HTML]{000000} 219} & {\color[HTML]{000000} 53} & {\color[HTML]{000000} -} & 119 & {\color[HTML]{000000} 68} & {\color[HTML]{000000} 47} & {\color[HTML]{000000} 376} & {\color[HTML]{000000} 170} & {\color[HTML]{000000} -} & {\color[HTML]{000000} 219} \\ \hline
\end{tabular}
}
\end{table}

\subsection{Ablation study}
The goal of these ablation experiments is to analyze the components in the pseudo-GT generation that effect the performance of 3SD. We perform seven experiments, patch-wise contrastive learning based self-supervised classification experiments (PCL16: $16\times 16$, PCL24: $24\times 24$, PCL32: $32\times 32$, PCL48: $48\times 48$ pixels), self-supervised classification with one global class label (GCL), PCM: generating pseudo-GT without using the edge map $e$ (using CM as $s_{pseudo}$), PGE: generating pseudo-GT without using the CAM map CM (using $g_e$ as $s_{pseudo}$), CMG: training 3SD with the pseudo-GT ($s_{pseudo} = CM \cup g_e$) and without $\mathcal{L}_{gs}$.

\noindent\textbf{Impact of patch-wise contrastive learning based self-supervised classification:} As can be seen from Table~\ref{tab:ablation}, PG (global class wise classification) fails to produce proper CAM maps (see Fig.~\ref{Fig:fig_cmaps}) which results in lower performance when compared to patch-wise contrastive learning based self-supervised classification with patch-size PCL16, PCL24, PCL32, PCL40 ($16\times 16,\: 24 \times24,\: 32\times 32,\: 48\times 48$ respectively). From Table~\ref{tab:ablation} and Fig.~\ref{Fig:fig_cmaps}, it is evident that when we increase the patch size in patch-wise self-supervised classification from $16\times 16$ to $\: 32\times 32$, we obtain better quality CAM maps which in turn results in better pseudo-GT and as a result better SOD performance. We obtained the best performance using setting of $32 \times 32$. Note that, larger patch doesn't improve the performance of 3SD method.

\begin{figure}[h!]
\centering
	\begin{center}
		\includegraphics[width=0.9\linewidth]{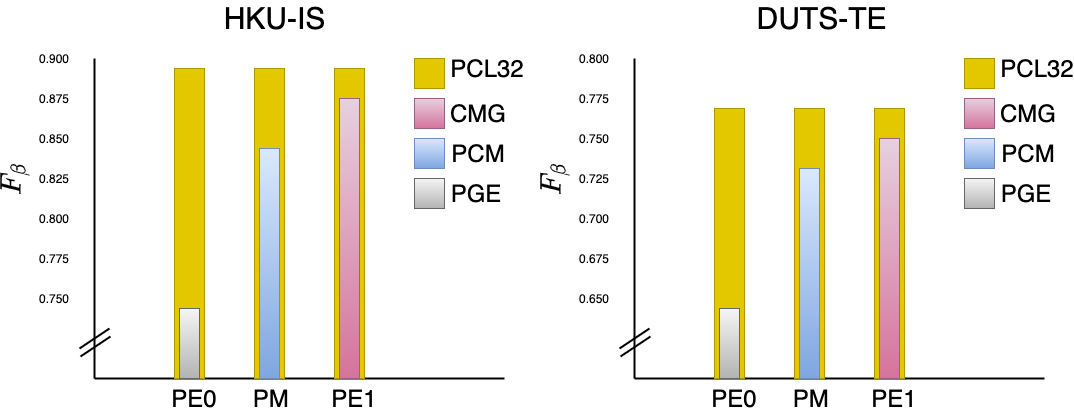} 
	\end{center}
	\vskip -17pt 
	\caption{Bar graph showing the improvements obtained when using $CM$ (CAM maps computed with help of self-supervised classification task) in the proposed pseudo-GT generation.}
	\label{fig:bargraph}
\end{figure}

\noindent\textbf{Impact of Pseudo-GT:} 
As explained in the section~\ref{sec:pseudo_gt},  pseudo-GT is defined as $s_{pseudo} = CM \cup g_e$. Here we perform experiments to validate the important role played by CAM ($CM$) map and gated edge ($g_e$) in the construction of the pseudo-GT. From Table~\ref{tab:ablation} columns PCM and PGE, we can  clearly observe a huge improvement in performance when we use $CM$ as the pseudo-GT instead of $g_e$. This shows that $CM$ obtained from patch-wise self-supervised classification contains more consistent semantic information than gated edge ($g_e$). Furthermore in the LPG column of Table~\ref{tab:ablation}, the  combination of both $CM$ and $g_e$ brings in further increase of the performance. The PCL32 column in Table~\ref{tab:ablation} corresponds to the case where we train 3SD with the pseudo-GT ($s_{pseudo} = CM \cup g_e$), with additional boundary aware loss ($\mathcal{L}_{gs}$).  In this case, we found a small improvement in the performance. Results corresponding to this ablation experiments are also shown in a bar-graph in Fig.~\ref{fig:bargraph}.

\noindent\textbf{BaseNetwork (BN):} We perform three experiments to evaluate the effectiveness of the constructed Base Network (BN). We use the following definitions in this experiment: 1) B0: BN with one encoder (local encoder $E_L$) and one decoder(saliency decoder $De_{S}$).  2) B1: BN with two encoders($E_L$ and $E_G$), and one decoder($De_{S}$) (adding global encoder ($E_G$) to BN). 3) B2: Using two encoders and two decoders (adding classification decoder to BN), as shown in Fig.~\ref{fig:overview}.  For this experiment, B0, B1, B2 methods are trained in supervised fashion using the actual ground-truth labels with cross-entropy loss. As can be seen from Table~\ref{tab:ablation}, we obtain an improvement when we add transformer based global encoder ($E_G$) to BN (B1 in Table~\ref{tab:ablation}).  This implies that $E_G$ is efficient in capturing patch-wise relations to obtain better saliency maps. Furthermore, adding a classification decoder improves BN's performance (B2 in Table~\ref{tab:ablation}). 
\vspace{-2em}

\vspace{-2em}
\begin{table}[h!]
\centering
\caption{Ablation study experiment on the DUTS-TE, HKU-IS, and ECSSD datasets. $\uparrow \& \downarrow$ denote larger and smaller is better, respectively.}
\label{tab:ablation}
\vskip-10pt
\resizebox{\linewidth}{!}{
\begin{tabular}{|l|l|ccccc|ccc|ccc|}
\hline
{} & {} & \multicolumn{8}{c|}{{Self or Unsupervised}} & \multicolumn{3}{c|}{{\begin{tabular}[c]{@{}c@{}}Fully-supervised \\ Base Network (BN)\end{tabular}}} \\ \cline{3-13}
\multirow{-2}{*}{{Dataset}} & \multirow{-2}{*}{{Metrics}} & {PCL16} & {PCL24} & {PCL32} & {PCL48}& {GCL} & {PCM} & {PGE} & {LPG} & {B0} & {B1} & {B2}  \\ \hline
{} & {$S_m\uparrow$} & {0.832} & {0.841} & {0.844} & {0.842} & \multicolumn{1}{l|}{{0.804}} & {0.791} & {0.712} & {0.828} & {0.836} & {0.861}  & {0.883} \\
{} & {$B_{\mu}\downarrow$} & {0.451} & {0.438} & {0.443} & {0.432} & \multicolumn{1}{l|}{{0.463}} & {0.621} & {0.658} & {0.501} & {0.368} & {0.364} & {0.358} \\
{} & {$F_{\beta}\uparrow$} & {0.759} & {0.762} & {0.765} & {0.760} & \multicolumn{1}{l|}{{0.748}} & {0.738} & {0.652} & {0.756} & {0.798} & {0.807} & {0.829} \\
{} & {$E_{\eta}\uparrow$} & {0.866} & {0.874} & {0.875} & {0.869} & \multicolumn{1}{l|}{{0.844}} & {0.832} & {0.740} & {0.858} & {0.884} & {0.897} & {0.913} \\
\multirow{-5}{*}{{DUTS-TE}} & {$MAE\downarrow$} & {0.047} & {0.044} & {0.044} & {0.052} & \multicolumn{1}{l|}{{0.058}} & {0.072} & {0.119} & {0.052} & {0.052} & {0.046}  & {0.036} \\ \hline
{} & {$S_m\uparrow$} & \multicolumn{1}{l}{{0.792}} & \multicolumn{1}{l}{{0.796}} & \multicolumn{1}{l}{{0.803}} & \multicolumn{1}{l}{{0.795}} & \multicolumn{1}{l|}{{0.782}} & {0.767} & \multicolumn{1}{l}{{0.672}} & \multicolumn{1}{l|}{{0.783}} & \multicolumn{1}{l}{{0.828}} & \multicolumn{1}{l}{{0.832}} &  \multicolumn{1}{l|}{{0.843}} \\
{} & {$B_{\mu}\downarrow$} & \multicolumn{1}{l}{{0.504}} & \multicolumn{1}{l}{{0.501}} & \multicolumn{1}{l}{{0.501}} & \multicolumn{1}{l}{{0.508}} & \multicolumn{1}{l|}{{0.511}} & {0.701} & \multicolumn{1}{l}{{0.687}} & \multicolumn{1}{l|}{{0.518}} & \multicolumn{1}{l}{{0.460}} & \multicolumn{1}{l}{{0.451}} & \multicolumn{1}{l|}{{0.429}} \\
{} & {$F_{\beta}\uparrow$} & \multicolumn{1}{l}{{0.726}} & \multicolumn{1}{l}{{0.738}} & \multicolumn{1}{l}{{0.735}} &  \multicolumn{1}{l}{{0.732}} & \multicolumn{1}{l|}{{0.715}} & {0.691} & \multicolumn{1}{l}{{0.614}} & \multicolumn{1}{l|}{{0.718}} & \multicolumn{1}{l}{{0.747}} & \multicolumn{1}{l}{{0.756}} &  \multicolumn{1}{l|}{{0.777}} \\
{} & {$E_{\eta}\uparrow$} & \multicolumn{1}{l}{{0.832}} & \multicolumn{1}{l}{{0.838}} & \multicolumn{1}{l}{{0.845}} & \multicolumn{1}{l}{{0.840}} & \multicolumn{1}{l|}{{0.826}} & {0.810} & \multicolumn{1}{l}{{0.731}} & \multicolumn{1}{l|}{{0.817}} & \multicolumn{1}{l}{{0.854}} & \multicolumn{1}{l}{{0.859}} &  \multicolumn{1}{l|}{{0.869}} \\
\multirow{-5}{*}{{DUT-OMRON}} & {$MAE\downarrow$} & \multicolumn{1}{l}{{0.062}} & \multicolumn{1}{l}{{0.062}} & \multicolumn{1}{l}{{0.060}} & \multicolumn{1}{l}{{0.062}} &\multicolumn{1}{l|}{{0.070}} & {0.077} & \multicolumn{1}{l}{{0.130}} & \multicolumn{1}{l|}{{0.063}} & \multicolumn{1}{l}{{0.060}} & \multicolumn{1}{l}{{0.054}} &  \multicolumn{1}{l|}{{0.049}} \\ \hline
{} & {$S_m\uparrow$} & {0.878} & {0.881} & {0.883} & {0.879} & {0.869} & {0.852} & {0.770} & {0.876} & {0.895} & {0.905} & {0.924} \\
{} & {$B_{\mu}\downarrow$} & {0.395} & {0.389} & {0.387} & {0.391} &{0.410} & {0.590} & {0.662} & {0.400} & {0.346} & {0.332}  & {0.319} \\
{} & {$F_{\beta}\uparrow$} & {0.881} & {0.891} & {0.897} & {0.894} &{0.873} & {0.843} & {0.756} & {0.892} & {0.891} & {0.905} & {0.933} \\
{} & {$E_{\eta}\uparrow$} & {0.921} & {0.926} & {0.928} & {0.928} & {0.904} & {0.869} & {0.854} & {0.922} & {0.929} & {0.934}  & {0.940} \\
\multirow{-5}{*}{{ECSSD}} & {$MAE\downarrow$} & {0.050} & {0.049} & {0.049} & {0.049} & {0.052} & {0.077} & {0.128} & {0.049} & {0.047} & {0.043}  & {0.032} \\ \hline
{} & {$S_m\uparrow$} & {0.862} & {0.891} & {0.904} & {0.896} & {0.863} & {0.841} & {0.747} & {0.892} & {0.890} & {0.902} & {0.918} \\
{} & {$B_{\mu}\downarrow$} & {0.389} & {0.384} & {0.381} & {0.383}  & {0.394} & {0.543} & {0.657} & {0.397} & {0.354} & {0.338}  & {0.321} \\
{} & {$F_{\beta}\uparrow$} & {0.864} & {0.881} & {0.885} & {0.883} & {0.859} & {0.849} & {0.734} & {0.878} & {0.893} & {0.903} & {0.918} \\
{} & {$E_{\eta}\uparrow$} & {0.920} & {0.928} & {0.935} & {0.932} & {0.916} & {0.901} & {0.829} & {0.931} & {0.941} & {0.946} & {0.959} \\
\multirow{-5}{*}{{HKU-IS}} & {$MAE\downarrow$} & {0.042} & {0.038} & {0.039} & {0.040}  & {0.048} & {0.055} & {0.125} & {0.040} & {0.035} & {0.031} & {0.024} \\ \hline
\end{tabular}
}
\end{table}

\vspace{-2em}
\section{Conclusion}
We presented a Self-Supervised Saliency Detection method (3SD), which doesn't require any labels, \textit{i.e.}, neither human annotations nor weak labels like image captions, handcrafted features or scribble annotations. The cornerstone of successful self-supervised SOD approach is generation of high-quality of pseudo-GTs. Our novel patch-wise contrastive learning paradigm effectively captures the semantics of salient objects. And this is the key of our superior performance verified on six benchmarks.

%
%
\bibliographystyle{splncs04}
\bibliography{egbib}

\begin{thebibliography}{10}
\providecommand{\url}[1]{\texttt{#1}}
\providecommand{\urlprefix}{URL }
\providecommand{\doi}[1]{https://doi.org/#1}

\bibitem{brendel2019approximating}
Brendel, W., Bethge, M.: Approximating cnns with bag-of-local-features models
  works surprisingly well on imagenet. arXiv preprint arXiv:1904.00760  (2019)

\bibitem{caron2020unsupervised}
Caron, M., Misra, I., Mairal, J., Goyal, P., Bojanowski, P., Joulin, A.:
  Unsupervised learning of visual features by contrasting cluster assignments.
  arXiv preprint arXiv:2006.09882  (2020)

\bibitem{caron2021emerging}
Caron, M., Touvron, H., Misra, I., J{\'e}gou, H., Mairal, J., Bojanowski, P.,
  Joulin, A.: Emerging properties in self-supervised vision transformers. arXiv
  preprint arXiv:2104.14294  (2021)

\bibitem{chen2017deeplab}
Chen, L.C., Papandreou, G., Kokkinos, I., Murphy, K., Yuille, A.L.: Deeplab:
  Semantic image segmentation with deep convolutional nets, atrous convolution,
  and fully connected crfs. TPAMI  \textbf{40}(4),  834--848 (2017)

\bibitem{chen2020simple}
Chen, T., Kornblith, S., Norouzi, M., Hinton, G.: A simple framework for
  contrastive learning of visual representations. In: ICML. pp. 1597--1607.
  PMLR (2020)

\bibitem{cheng2014salientshape}
Cheng, M.M., Mitra, N.J., Huang, X., Hu, S.M.: Salientshape: group saliency in
  image collections. The visual computer  \textbf{30}(4),  443--453 (2014)

\bibitem{dai2015boxsup}
Dai, J., He, K., Sun, J.: Boxsup: Exploiting bounding boxes to supervise
  convolutional networks for semantic segmentation. In: ICCV. pp. 1635--1643
  (2015)

\bibitem{dosovitskiy2020image}
Dosovitskiy, A., Beyer, L., Kolesnikov, A., Weissenborn, D., Zhai, X.,
  Unterthiner, T., Dehghani, M., Minderer, M., Heigold, G., Gelly, S., et~al.:
  An image is worth 16x16 words: Transformers for image recognition at scale.
  arXiv preprint arXiv:2010.11929  (2020)

\bibitem{dosovitskiy2015discriminative}
Dosovitskiy, A., Fischer, P., Springenberg, J.T., Riedmiller, M., Brox, T.:
  Discriminative unsupervised feature learning with exemplar convolutional
  neural networks. TPAMI  \textbf{38}(9),  1734--1747 (2015)

\bibitem{fan2017structure}
Fan, D.P., Cheng, M.M., Liu, Y., Li, T., Borji, A.: Structure-measure: A new
  way to evaluate foreground maps. In: ICCV. pp. 4548--4557 (2017)

\bibitem{fan2018enhanced}
Fan, D.P., Gong, C., Cao, Y., Ren, B., Cheng, M.M., Borji, A.:
  Enhanced-alignment measure for binary foreground map evaluation. arXiv
  preprint arXiv:1805.10421  (2018)

\bibitem{feng2019attentive}
Feng, M., Lu, H., Ding, E.: Attentive feedback network for boundary-aware
  salient object detection. In: CVPR. pp. 1623--1632 (2019)

\bibitem{grill2020bootstrap}
Grill, J.B., Strub, F., Altché, F., Tallec, C., Richemond, P.H., Buchatskaya,
  E., Doersch, C., Pires, B.A., Guo, Z.D., Azar, M.G., Piot, B., Kavukcuoglu,
  K., Munos, R., Valko, M.: Bootstrap your own latent: A new approach to
  self-supervised learning. NIPS  (2020)

\bibitem{gutmann2010noise}
Gutmann, M., Hyv{\"a}rinen, A.: Noise-contrastive estimation: A new estimation
  principle for unnormalized statistical models. In: Proceedings of the
  thirteenth international conference on artificial intelligence and
  statistics. pp. 297--304 (2010)

\bibitem{he2020momentum}
He, K., Fan, H., Wu, Y., Xie, S., Girshick, R.: Momentum contrast for
  unsupervised visual representation learning. In: CVPR. pp. 9729--9738 (2020)

\bibitem{henaff2020data}
Henaff, O.: Data-efficient image recognition with contrastive predictive
  coding. In: ICML. pp. 4182--4192. PMLR (2020)

\bibitem{hermans2017defense}
Hermans, A., Beyer, L., Leibe, B.: In defense of the triplet loss for person
  re-identification. arXiv preprint arXiv:1703.07737  (2017)

\bibitem{hsu122017weakly}
Hsu12, K.J., Lin, Y.Y., Chuang, Y.Y.: Weakly supervised saliency detection with
  a category-driven map generator. BMVC  (2017)

\bibitem{khoreva2017simple}
Khoreva, A., Benenson, R., Hosang, J., Hein, M., Schiele, B.: Simple does it:
  Weakly supervised instance and semantic segmentation. In: CVPR. pp. 876--885
  (2017)

\bibitem{kim2014salient}
Kim, J., Han, D., Tai, Y.W., Kim, J.: Salient region detection via
  high-dimensional color transform. In: CVPR. pp. 883--890 (2014)

\bibitem{li2018weakly}
Li, G., Xie, Y., Lin, L.: Weakly supervised salient object detection using
  image labels. In: AAAI. pp. 7024--7031 (2018)

\bibitem{li2015visual}
Li, G., Yu, Y.: Visual saliency based on multiscale deep features. In: CVPR.
  pp. 5455--5463 (2015)

\bibitem{li2014secrets}
Li, Y., Hou, X., Koch, C., Rehg, J.M., Yuille, A.L.: The secrets of salient
  object segmentation. In: CVPR. pp. 280--287 (2014)

\bibitem{liu2018picanet}
Liu, N., Han, J., Yang, M.H.: Picanet: Learning pixel-wise contextual attention
  for saliency detection. In: CVPR. pp. 3089--3098 (2018)

\bibitem{liu2021visual}
Liu, N., Zhang, N., Wan, K., Shao, L., Han, J.: Visual saliency transformer.
  In: CVPR. pp. 4722--4732 (2021)

\bibitem{liu2017richer}
Liu, Y., Cheng, M.M., Hu, X., Wang, K., Bai, X.: Richer convolutional features
  for edge detection. In: CVPR. pp. 3000--3009 (2017)

\bibitem{long2015fully}
Long, J., Shelhamer, E., Darrell, T.: Fully convolutional networks for semantic
  segmentation. In: CVPR. pp. 3431--3440 (2015)

\bibitem{lu2013robust}
Lu, S., Tan, C., Lim, J.H.: Robust and efficient saliency modeling from image
  co-occurrence histograms. TPAMI  \textbf{36}(1),  195--201 (2013)

\bibitem{luo2017non}
Luo, Z., Mishra, A., Achkar, A., Eichel, J., Li, S., Jodoin, P.M.: Non-local
  deep features for salient object detection. In: CVPR. pp. 6609--6617 (2017)

\bibitem{nguyen2019deepusps}
Nguyen, D.T., Dax, M., Mummadi, C.K., Ngo, T.P.N., Nguyen, T.H.P., Lou, Z.,
  Brox, T.: Deepusps: Deep robust unsupervised saliency prediction with
  self-supervision. arXiv preprint arXiv:1909.13055  (2019)

\bibitem{obukhov2019gated}
Obukhov, A., Georgoulis, S., Dai, D., Van~Gool, L.: Gated crf loss for weakly
  supervised semantic image segmentation. arXiv preprint arXiv:1906.04651
  (2019)

\bibitem{Piao_2021_ICCV}
Piao, Y., Wang, J., Zhang, M., Lu, H.: Mfnet: Multi-filter directive network
  for weakly supervised salient object detection. In: ICCV. pp. 4136--4145
  (October 2021)

\bibitem{qin2020u2}
Qin, X., Zhang, Z., Huang, C., Dehghan, M., Zaiane, O.R., Jagersand, M.:
  U2-net: Going deeper with nested u-structure for salient object detection.
  Pattern Recognition  \textbf{106},  107404 (2020)

\bibitem{Qin_2019_CVPR}
Qin, X., Zhang, Z., Huang, C., Gao, C., Dehghan, M., Jagersand, M.: Basnet:
  Boundary-aware salient object detection. In: CVPR. pp. 7479--7489 (June 2019)

\bibitem{sohn2016improved}
Sohn, K.: Improved deep metric learning with multi-class n-pair loss objective.
  NIPS  \textbf{29} (2016)

\bibitem{tarvainen2017mean}
Tarvainen, A., Valpola, H.: Mean teachers are better role models:
  Weight-averaged consistency targets improve semi-supervised deep learning
  results. arXiv preprint arXiv:1703.01780  (2017)

\bibitem{tian2020contrastive}
Tian, Y., Krishnan, D., Isola, P.: Contrastive multiview coding. In: ECCV. pp.
  776--794. Springer (2020)

\bibitem{wang2017learning}
Wang, L., Lu, H., Wang, Y., Feng, M., Wang, D., Yin, B., Ruan, X.: Learning to
  detect salient objects with image-level supervision. In: CVPR. pp. 136--145
  (2017)

\bibitem{wang2018detect}
Wang, T., Zhang, L., Wang, S., Lu, H., Yang, G., Ruan, X., Borji, A.: Detect
  globally, refine locally: A novel approach to saliency detection. In: CVPR.
  pp. 3127--3135 (2018)

\bibitem{CVPR2020_LDF}
Wei, J., Wang, S., Wu, Z., Su, C., Huang, Q., Tian, Q.: Label decoupling
  framework for salient object detection. In: CVPR. pp. 13022--13031 (June
  2020)

\bibitem{wu2019mutual}
Wu, R., Feng, M., Guan, W., Wang, D., Lu, H., Ding, E.: A mutual learning
  method for salient object detection with intertwined multi-supervision. In:
  CVPR. pp. 8150--8159 (2019)

\bibitem{wu2019cascaded}
Wu, Z., Su, L., Huang, Q.: Cascaded partial decoder for fast and accurate
  salient object detection. In: CVPR. pp. 3907--3916 (2019)

\bibitem{wu2018unsupervised}
Wu, Z., Xiong, Y., Yu, S.X., Lin, D.: Unsupervised feature learning via
  non-parametric instance discrimination. In: CVPR. pp. 3733--3742 (2018)

\bibitem{yan2013hierarchical}
Yan, Q., Xu, L., Shi, J., Jia, J.: Hierarchical saliency detection. In: CVPR.
  pp. 1155--1162 (2013)

\bibitem{yang2013saliency}
Yang, C., Zhang, L., Lu, H., Ruan, X., Yang, M.H.: Saliency detection via
  graph-based manifold ranking. In: CVPR. pp. 3166--3173 (2013)

\bibitem{zeng2019multi}
Zeng, Y., Zhuge, Y., Lu, H., Zhang, L., Qian, M., Yu, Y.: Multi-source weak
  supervision for saliency detection. In: CVPR. pp. 6074--6083 (2019)

\bibitem{zhang2017supervision}
Zhang, D., Han, J., Zhang, Y.: Supervision by fusion: Towards unsupervised
  learning of deep salient object detector. In: ICCV. pp. 4048--4056 (2017)

\bibitem{zhang2020learning}
Zhang, J., Xie, J., Barnes, N.: Learning noise-aware encoder-decoder from noisy
  labels by alternating back-propagation for saliency detection. In: ECCV. pp.
  349--366. Springer (2020)

\bibitem{zhang2020weakly}
Zhang, J., Yu, X., Li, A., Song, P., Liu, B., Dai, Y.: Weakly-supervised
  salient object detection via scribble annotations. In: CVPR. pp. 12546--12555
  (2020)

\bibitem{zhang2018deep}
Zhang, J., Zhang, T., Dai, Y., Harandi, M., Hartley, R.: Deep unsupervised
  saliency detection: A multiple noisy labeling perspective. In: CVPR. pp.
  9029--9038 (2018)

\bibitem{zhang2017novel}
Zhang, J., Ehinger, K.A., Wei, H., Zhang, K., Yang, J.: A novel graph-based
  optimization framework for salient object detection. Pattern Recognition
  \textbf{64},  39--50 (2017)

\bibitem{zhang2017learning}
Zhang, P., Wang, D., Lu, H., Wang, H., Yin, B.: Learning uncertain
  convolutional features for accurate saliency detection. In: ICCV. pp.
  212--221 (2017)

\bibitem{zhang2019salient}
Zhang, Q., Huo, Z., Liu, Y., Pan, Y., Shan, C., Han, J.: Salient object
  detection employing a local tree-structured low-rank representation and
  foreground consistency. Pattern Recognition  \textbf{92},  119--134 (2019)

\bibitem{zhang2018progressive}
Zhang, X., Wang, T., Qi, J., Lu, H., Wang, G.: Progressive attention guided
  recurrent network for salient object detection. In: CVPR. pp. 714--722 (2018)

\bibitem{zhou2016learning}
Zhou, B., Khosla, A., Lapedriza, A., Oliva, A., Torralba, A.: Learning deep
  features for discriminative localization. In: CVPR. pp. 2921--2929 (2016)

\bibitem{zhu2014saliency}
Zhu, W., Liang, S., Wei, Y., Sun, J.: Saliency optimization from robust
  background detection. In: CVPR. pp. 2814--2821 (2014)

\end{thebibliography}

\end{document}